\documentclass[sigconf]{acmart}

\AtBeginDocument{%
  }

\setcopyright{acmcopyright}
\copyrightyear{2023} 
\acmYear{2023} 
\setcopyright{acmlicensed}\acmConference[WWW '23]{Proceedings of the ACM Web Conference 2023}{May 1--5, 2023}{Austin, TX, USA}
\acmBooktitle{Proceedings of the ACM Web Conference 2023 (WWW '23), May 1--5, 2023, Austin, TX, USA}
\acmPrice{15.00}
\acmDOI{10.1145/3543507.3583344}
\acmISBN{978-1-4503-9416-1/23/04}

\usepackage{algorithm}
\usepackage{algorithmic}
\usepackage{newfloat}
\usepackage{listings}
\usepackage{amsmath}
\usepackage{amsfonts}
\usepackage{amssymb}
\usepackage{amsthm}
\usepackage{subfigure}
\usepackage{float}
\usepackage{multirow}
\usepackage{multicol}
\usepackage{booktabs}
\usepackage{enumitem}
\usepackage{bm}
\usepackage[normalem]{ulem}

\newtheorem{theorem}{Theorem}[section]

\newtheorem{prop}[theorem]{Proposition}

\newtheorem{remark}{Remark}

\def \mL{\mathcal L}

\newcommand{\cU}{\mathcal{U}}
\newcommand{\cV}{\mathcal{V}}
\newcommand{\cS}{\mathcal{S}}
\newcommand{\cQ}{\mathcal{Q}}
\newcommand{\cT}{\mathcal{T}}

\usepackage{array}
\newcolumntype{L}[1]{>{\raggedright\let\newline\\\arraybackslash\hspace{0pt}}m{#1}}
\newcolumntype{C}[1]{>{\centering\let\newline  \\\arraybackslash\hspace{0pt}}m{#1}}
\newcolumntype{R}[1]{>{\raggedleft\let\newline \\\arraybackslash\hspace{0pt}}m{#1}}

\usepackage{xcolor}
\newcommand{\TheName}[0]{ColdNAS}

\begin{document}

\fancyhead{}

\title{ColdNAS: Search to Modulate for User Cold-Start Recommendation}


\author{Shiguang Wu}
\affiliation{%
	\institution{Department of Electronic Engineering\\ Tsinghua University}
 \city{Beijing}
 \country{China}}
	\email{wsg19@mails.tsinghua.edu.cn}

\author{Yaqing Wang}
\authornote{Corresponding author.}
\affiliation{%
	\institution{Baidu Inc.}
	 \city{Beijing}
	\country{China}}
	\email{wangyaqing01@baidu.com}

\author{Qinghe Jing}
\affiliation{%
	\institution{Baidu Inc.}
 \city{Beijing}
\country{China}}
	\email{jingqinghe@baidu.com}

\author{Daxiang Dong}
\affiliation{%
	\institution{Baidu Inc.}
	 \city{Beijing}
	\country{China}}
	\email{dongdaxiang@baidu.com}

\author{Dejing Dou}
\affiliation{%
	\institution{Baidu Inc.}
	 \city{Beijing}
	\country{China}}
	\email{doudejing@baidu.com}

\author{Quanming Yao}
\affiliation{%
	\institution{Department of Electronic Engineering\\ Tsinghua University}
	\city{Beijing}
	\country{China}}
\email{qyaoaa@tsinghua.edu.cn}


\begin{abstract}
Making personalized recommendation for cold-start users, who only have a few interaction histories, is a challenging problem in recommendation systems. 
Recent works leverage hypernetworks to directly map user interaction histories to user-specific parameters, which are then used to modulate predictor by feature-wise linear modulation function. These works obtain the state-of-the-art performance. However, the physical meaning of scaling and shifting in recommendation data is unclear. Instead of using a fixed modulation function and deciding modulation position by expertise, we propose a modulation framework called ColdNAS for user cold-start problem, where we look for proper modulation structure, including function and position, via neural architecture search. 
We design a search space which covers broad models and theoretically prove that this search space can be transformed to a much smaller space, enabling an efficient and robust one-shot search algorithm.
Extensive experimental results on benchmark datasets show that ColdNAS consistently performs the best. We observe that different modulation functions lead to the best performance on different datasets, which validates the necessity of designing a searching-based method. 
Codes are available at 
\url{https://github.com/LARS-research/ColdNAS}. 
\end{abstract}

\maketitle

\section{Introduction}
\label{sec:intro}

%
%

Recommendation systems (RSs)~\cite{resnick1997recommender}
target at
providing suggestions of items that are most pertinent to a particular user, such as  movie recommendation~\cite{harper2015movielens} and 
book recommendation~\cite{ziegler2005improving}. 
Nowadays, RSs are abundant online, offering enormous users convenient ways to shopping regardless of location and time, and also providing intimate suggestions according to their preferences. 
However, user cold-start recommendation \cite{schein2002methods} remains a severe problem in RSs. 
On the one hand, the users in RSs follow long tail effect \cite{park2008long}, some users just have a few interaction histories. On the other hand, new users are continuously emerging, who naturally have rated a few items in RSs. 
Such
a problem
is even more challenging 
as modern RSs are mostly built with over-parameterized
deep networks,
which needs a huge amount of training samples to get good performance
and can easily overfit for cold-start users~\cite{volkovs2017dropoutnet}. 

\begin{figure}[t]
	\centering
	\vspace{0px}
	\includegraphics[width=0.40\textwidth]{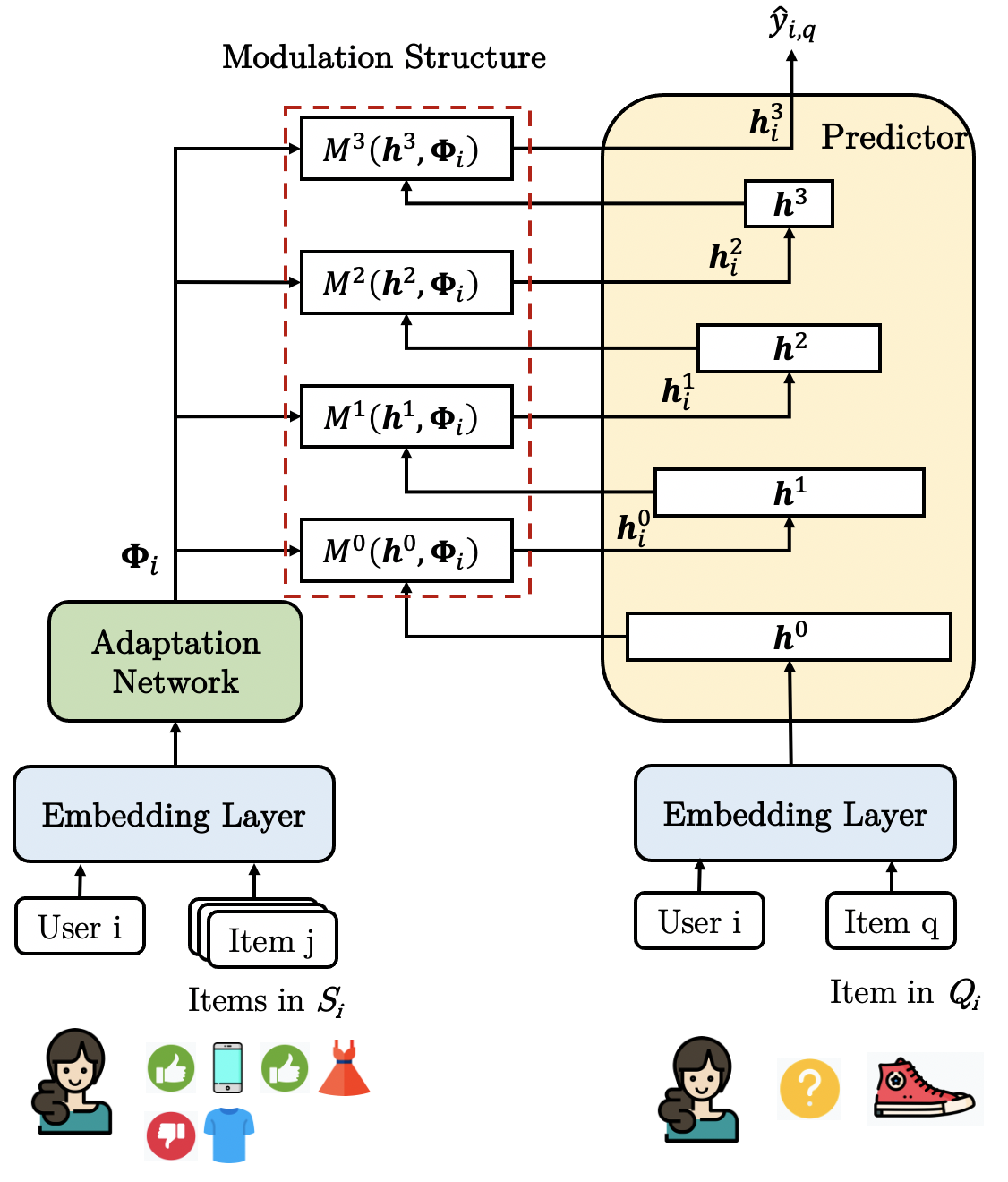}
	
	\vspace{-5px}
	\caption{Architecture used in user cold-start models, which consists of embedding layer, 
		adaptation network and predictor. Modulation structure in red dotted lines are the searching target of our paper.}
	\label{fig:mod_framework}
	\vspace{-15px}
\end{figure}

User cold-start recommendation problem can naturally be modeled as a few-shot learning problem \cite{wang2020generalizing},  which targets at quickly generalize to new tasks (i.e. personalized recommendation for cold-start users) with a few training samples (i.e. a few interaction histories). 
A number of works~\cite{lee2019melu,dong2020mamo,lu2020meta,yu2021personalized,wang2021preference} 
adopt the classic gradient-based meta-learning strategy called 
model-agnostic meta-learning (MAML) \citep{finn2017model}, which learns a good initialized parameter from a set of tasks and adapts it to 
a new task by taking a few steps of gradient descent updates on a limited number of labeled samples.  
This line of models has demonstrated high potential of alleviating user cold-start problem. 
However, gradient-based meta-learning strategy require expertise to tune the optimization procedure to avoid over-fitting. Besides, the inference time can be long. 

Instead of adapting to each user by fine-tuning via gradient descent,
another line of works 
uses hypernetworks \cite{ha2016hypernetworks} to directly map user interaction history to user-specific parameters~\cite{dong2020mamo,lin2021task,feng2021cmml,pang2022pnmta}. These modulation-based methods consist of embedding layer,
adaptation network
and 
predictor. The adaptation network generates user-specific parameters, which are used to modulate the predictor in the form of a modulation function. 
Particularly, 
they all adopt 
feature-wise linear modulation function (FiLM)~\cite{perez2018film}, 
which modulates the representation via scaling and shifting based on the conditioning information, to modulate the user cold-start models to obtain user-specific representation. 
Although FiLM 
has been proved to be highly effective in
on images~\cite{requeima2019fast} and graphs such molecules and protein-protein interaction graphs~\cite{brockschmidt2020gnn}, 
applying scaling and shifting on user interaction history has rather obscure physical meaning.  Moreover, choosing where to modulate is hard to decide. Existing works modulate different parts of the model, such as last layers of the decoder~\cite{lin2021task} and most layers in the model~\cite{pang2022pnmta}. 
How to modulate well for different users, and how to choose the right functions at the right positions to modulate, are still open questions. 


In this paper, we propose ColdNAS to find appropriate modulation structure for user cold-start problem by 
neural architecture search~(NAS). 
Although NAS methods have been applied in RSs ~\cite{gao2021progressive,xie2021fives}, the design of NAS methods are problem-specific. For user cold-start problem, it is still unknown 
how to (i) design a search space that can cover effective cold-start models with good performance for various datasets, and (ii) design an efficient and robust search algorithm. 
To solve the above challenges, we design a search space of modulation structure, which can cover not only existing modulation-based user cold-start models, but also contain more flexible structures.
We theoretically prove that the proposed search space can be transformed to an equivalent space, where we search efficiently and robustly by differentiable architecture search. 
Our main contributions are summarized as follows: 
\begin{itemize}[leftmargin=*]
\item 
We propose \TheName{}, a modulation framework  for user cold-start problem. 
We use a hypernetwork to map each user's history interactions to user-specific parameters which are then used to modulate the predictor, and formulate how to modulate and where to modulate
as a NAS problem. 

\item 
We design a search space of modulation structure, 
which can cover not only existing modulation-based user cold-start models, but also contain more expressive structures.  
As this search space can be large to search, we conduct search space transformation to transform the original space to an equivalent but much smaller space. Theoretical analysis is provided to validate its correctness. 
Upon the transformed space, 
we then can search efficiently and robustly by differentiable architecture search algorithm.  

\item We perform
extensive experiments on benchmark datasets for user cold-start problem, and observe that \TheName{} consistently obtains the state-of-the-art performance. 
We also validate the design consideration of search spaces and algorithms, demonstrating the strength and reasonableness of ColdNAS.
\end{itemize}

\section{Related Works}
\label{sec:related}

\subsection{User Cold-Start Recommendation}
\label{sec:rel_cold}
Making personalized recommendation for cold-start users is particular challenging, as these users only have a few interaction histories \cite{schein2002methods}.  
In the past, collaborative filtering (CF)-based  \citep{koren2008factorization,sedhain2015autorec,he2017neural} methods, which make predictions by capturing interactions among users and items to represent them in low-dimensional space, obtain leading performance in RSs. 
However, these CF-based methods make inferences only based on the user's history.   
They cannot handle user cold-start problem.  
To alleviate the user cold-start problem,  content-based methods 
leverage user/item features ~\cite{schein2002methods}  or even
user social relations~\cite{lin2013addressing} to help to predict for cold-start users.
The recent deep model DropoutNet \cite{volkovs2017dropoutnet} trains a neural network with dropout mechanism applied on input samples and infers the missing data. However, it is hard to generalize these content-based methods to new users, which usually requires model retraining. 

A recent trend is to model user cold-start problem as a few-shot learning problem~\cite{wang2020generalizing}.
The resultant models learn the ability to 
quickly generalize to recommend for new users 
with a few interaction histories. 
Most works  mainly 
follow the classic gradient-based meta-learning strategy MAML \citep{finn2017model}, 
which first learns a good initialized parameter from training tasks, then locally update the parameter on 
the provided interaction history by gradient descent. 
In particular, 
existing 
works consider different directions to improve the performance: 
MeLU~\cite{lee2019melu} selectively adapts model parameters to the new task in the local update stage,  
MAMO~\cite{dong2020mamo} introduces external memory to guide the model to adapt, 
MetaHIN~\cite{lu2020meta} uses heterogeneous information networks  to leverage the rich semantics between users and items, 
REG-PAML~\cite{yu2021personalized} proposes to use user-specific learning rate during local update, and 
PAML~\cite{wang2021preference}  leverages social relations to share information among similar users. 
While these approaches can adapt models to training data, they are computationally inefficient at test-time, and usually require expertise to tune the optimization procedure to avoid over-fitting.

\subsection{Neural Architecture Search}
\label{sec:rel_nas}

Neural architecture search (NAS) targets at finding an architecture with good performance without human tuning~\cite{hutter2019automated}. 
Recently, NAS methods have been applied in RSs.  
SIF~\cite{yao2020sif} searches for interaction function in collaborative filtering,   
AutoCF~\cite{gao2021efficient} further searches for basic components including input encoding, embedding function, interaction function,
and prediction function in collaborative filtering. 
AutoFIS~\cite{liu2020autofis}, AutoCTR~\cite{song2020towards} and FIVES~\cite{xie2021fives} search for effective feature interaction in click-through rate prediction. 
AutoLoss~\cite{zhao2021autoloss} searches for loss function in RSs.   
Due to different problem settings, search spaces needs to problem-specific and cannot be shared or transferred.  Therefore, 
none of these works can be applied for user cold-start recommendation problem.
To search efficiently on the search space, one can choose
 reinforcement learning methods~\cite{baker2016designing}, 
evolutionary algorithms~\cite{real2019regularized}, and one-shot differentiable architecture search algorithms \cite{liu2018darts,pham2018efficient,yao2020efficient}. 
Among them,  one-shot differentiable architecture search algorithms have demonstrated higher efficiency. 
Instead of training and evaluating different models like classical methods, 
they optimize only one supernet where the model parameters are shared across the search space and co-adapted.  

\section{Proposed Method}
\label{sec:proposed}

In this section, 
we present the details of \TheName{}, whose overall architecture is shown in  Figure~\ref{fig:mod_framework}. 
In the sequel, we first provide the formal problem formulation of user cold-start problem (Section~\ref{sec:proposed-problem}). 
Then, we present our search space and theoretically show how to transform it (Section~\ref{sec:proposed-space}).
Finally, we introduce the search algorithm to search for a good user cold-start model (Section~\ref{sec:proposed-alg}).

\subsection{Problem Formulation}
\label{sec:proposed-problem}
Let user set be denoted $\cU = \{ u_i \}$, where each user $u_i$ is associated with user features. 
The feature space is shared across all users. 
Let item set be denoted $\cV = \{ v_j \}$ where each item  $v_j$ is also associated with item features. 
When a user $u_i$ rates an item $v_j$, the rating is denoted $y_{i,j}$. 
In user cold-start recommendation problem, the focus is to make personalized recommendation for user $u_i$ who only has rated a few items. 

Following recent works \cite{bharadhwaj2019meta,lu2020meta,lin2021task}, we model the user cold-start recommendation problem as a few-shot learning problem. 
The target is to learn a model from a set of training user cold-start tasks $\cT^{\text{train}}$ and generalize to provide personalized recommendation for new tasks. 
Each task $T_i$ corresponds to a user 
$u_i$, 
with a support set $\cS_i = \{ (v_j, y_{i,j})  \}_{ j = 1 }^{N}$ containing existing interaction histories 
and a query set $\cQ_i = \{( v_{j},y_{i,j}) \}_{j=1}^{M}$ containing interactions to predict. 
$N$ and $M$ are the number of interactions in $\cS_i$. $\cQ_i$ and $N$ are small.

\subsection{Search Space}
\label{sec:proposed-space}

Existing modulation-based user cold-start works can be summarized into the following modulation framework 
consisting of three parts, 
i.e.,
embedding layer,
adaptation network
and 
predictor, 
as plotted in Figure~\ref{fig:mod_framework}.
\begin{itemize}[leftmargin=*]
	\item The embedding layer $E$ with parameter $\bm{\theta}_E$ 
	embeds the categorical features from users and items into dense vectors,
	i.e.,
	$(\bm{u}_i, \bm{v}_j) = E(u_i, v_j; \bm{\theta}_E)$.
	
	\item
	The adaptation network $A$ with parameter $\bm{\theta}_{A}$
	takes the support set $\cS_i$ for a specific user $u_i$ as input and
	generates user-specific adaptive parameters,
	i.e., 
	\begin{align}
		\bm{\Phi}_i =\{ \bm{\phi}_i^k \}_{ k= 1 }^{C}=A(\cS_i;\bm{\theta}_{A}),
		\label{eq:adapara}
	\end{align}
	where $C$ is the number of adaptive parameter groups for certain modulation structure.

	\item 
	The predictor $P$ with parameter $\bm{\theta}_P$ 
	takes user-specific parameters $\bm{\phi}_i$ 
	and $v_q\in \cQ_i$ from the query set 
	as input,
	and generate predictions by
	\begin{equation}
		\label{eq:forward}
		\hat{y}_{i,j} = P( (\bm{u}_i, \bm{v}_q), \bm{\Phi}_i; \bm{\theta}_{P}).
	\end{equation}
\end{itemize}

Comparing with
classical RS models,
the extra adaptation network is 
introduced to handle cold-start users. 
To make personalized recommendation, for each $u_i$, the support set $\cS_i$ is first mapped to user-specific parameter $\bm{\phi}_i$ by \eqref{eq:adapara}. Then taking the features of target item $v_q\in \cQ_i$, user features $u_i$, and the $\bm{\phi}_i$, prediction is made as \eqref{eq:forward}.
Subsequently,
how  to use the user-specific parameter $\bm{\phi}_i$  
to change the prediction process in \eqref{eq:forward}
can be crucial to the performance. 
Usually,
a multi-layer perception (MLP)
is used as $P$
\cite{cheng2016wide,he2017neural,lin2021task}.
Assume a $L$-layer MLP is used 
and $\bm{h}^l$ denotes its output from the $l$th layer, and let $\bm{h}^0= (\bm{u}_i, \bm{v}_q)$ for notation simplicity.
For the $i$th user, 
$\bm{h}^{l + 1}$ is modulated as 
\begin{align}
	\bm{h}_i^l
	& = M^l(\bm{h}^l,  \bm{\Phi}_i ),
	\label{eq:base-mod}
	\\
	\bm{h}^{l+1}
	& = \text{ReLU}( \bm{W}^l_P \bm{h}_i^l + \bm{b}^l_P),
	\label{eq:base-fc}
\end{align} 
where 
$M^l$ is the modulation function,
$\bm{W}^l$ and $\bm{b}^l$ are learnable weights at the $l$th layer.
This 
$M^l$ controls how 
$\bm{h}^l$ is personalized w.r.t. the $i$th user.
The recent TaNP~\cite{lin2021task} directly lets 
$M^l$ in \eqref{eq:base-mod} adopt the form of FiLM for all $L$ MLP layers: 
\begin{align}
	\bm{h}^l_i = \bm{h}^l \odot \bm{\phi}_i^1 + \bm{\phi}_i^2. 
	\label{eq:film}
\end{align}
FiLM applies a feature-wise affine transformation on the
intermediate features, and has been proved to be highly effective in
other domains~\cite{requeima2019fast,brockschmidt2020gnn}. 
However, 
users and items can have diverse interaction patterns.
For example,
both the inner product and summation have been used in RS
to measure the preference of user over items~\cite{koren2008factorization,hsieh2017collaborative}.

In order to find the appropriate modulation function, we are motivated to search $M^l$ for 
different recommendation tasks. 
We design the following space for $M^l$: 
\begin{align}
	\bm{h}^l_i 
	= \bm{h}^l \circ_{\text{op}^1} \bm{\phi}_i^1 \circ_{\text{op}^2} \bm{\phi}_i^2\cdots \circ_{\text{op}^C} \bm{\phi}_i^C ,
	\label{eq:fullsp}
\end{align}
where $\text{op}^i$' are defined as 
\begin{align*}
	\circ_{\text{op}_i} \in	
	\mathcal{O}
	\equiv
	\big\{
	\max, \min, \odot, /,  +,  - 
	\big\}. 
\end{align*}
They are all commonly used simple dimension-preserving binary operations. We choose them to avoid the resultant 
$M^l$ being too complex, which can easily overfit for 
cold-start users. 

\begin{figure*}[htbp]
	\centering
	\vspace{-10px}
	\subfigure[Original search space.]{
		\includegraphics[width=0.21\textwidth]{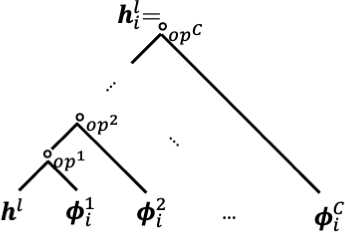}
		\label{fig:originsp}}
	\subfigure[Transformed search space.]{
		\includegraphics[width=0.2\textwidth]{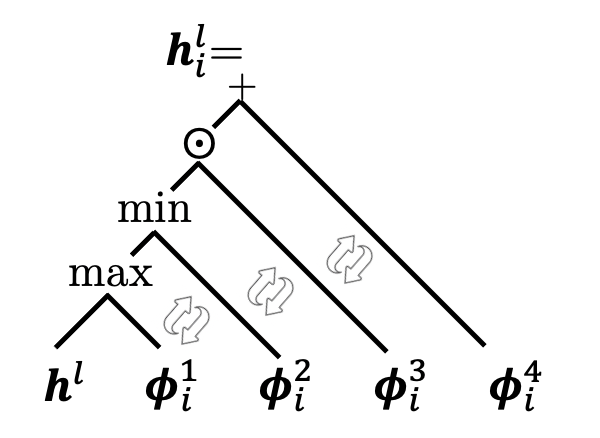}
		\label{fig:reducesp}}
	\subfigure[A layer in supernet.]{
		\includegraphics[width=0.48\textwidth]{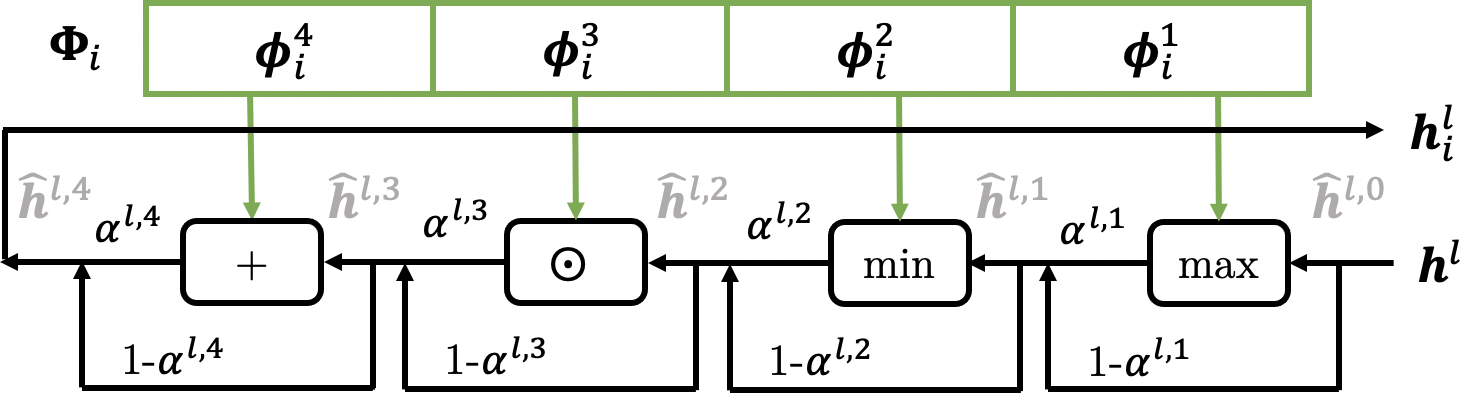}}
	\vspace{-10px}
	\caption{Illustration of our proposition~\ref{prop:space}, and the structure of the supernet to search on the reduced space.}
	\label{fig:space}
\end{figure*}

The search space in \eqref{eq:fullsp} can be viewed as a binary tree, as shown in Figure \ref{fig:originsp}.
Since $M^l$ can be different for each layer,
the size of this space is $6^{C\times L}$. 
A larger $C$ leads to a larger search space, which has higher potential of containing appropriate modulation function but is also more challenging to search effectively.  

\subsection{Search Strategy}
\label{sec:proposed-alg}

We aim to find an efficient and practicable search algorithm on the proposed search space.
However, the search space can be very large, and
differentiable NAS methods are known to be fragile
on large spaces~\citep{yu2019evaluating}.
In the sequel,
we propose to first transform the original search space to an equivalent but much smaller space, as
shown in Figure~\ref{fig:originsp}, where the equivalence is inspired by some similarities between the operations and the expressiveness of deep neural networks. On the transformed space, we design a supernet structure to conduct efficient and robust differentiable search.

\subsubsection{Search Space Transformation}
Though the aforementioned search space can be very large, 
we can transform it to an equivalent space of size $2^{4\times L}$ which is invariant with $C$, as proved in Proposition~\ref{prop:space}\footnote{The proof is in Appendix~\ref{app:proof}.} below.

\begin{prop}[Search Space Transformation] 
	\label{prop:space}
	Assume the adaptation network $A$ is expressive enough. 
	Any $M^l$ with a form of  \eqref{eq:fullsp}
	where $\circ_{\text{op}^k}
	\in
	\mathcal{O}$, 
	$C$ is any non-negative integer, and $ \bm{\phi}_i^k\in \bm{\Phi}_i =A(\cS_i,\bm{\theta}_{A})$, 
	can be represented as
	\begin{align}
		\label{eq:reduced}
		\bm{h}^l_i 
		=\min( \max(\bm{h}^l,~\hat{\bm{\phi}}_i^1),~\hat{\bm{\phi}}_i^2)\odot\hat{\bm{\phi}}_i^3 +\hat{\bm{\phi}}_i^4,
	\end{align}
	and the above four operations are permutation-invariant.
\end{prop}

The intuitions are as follows. 
First,
operations in $\mathcal{O}$ can be divided into 
four groups: $G_1=\{\max\},~G_2=\{\min\},~G_3=\{+,-\}, ~G_4=\{\odot,/\}$.
Then, with mild assumption on adaptation network, we can prove two important properties: 
(i) inner-group consistence: operations that in the same group can be associated; and 
(ii) inter-group permutation-invariance:  
operations that are not in the same group are permutation-invariant which means any two operations can switch with another. 
Thus, we can recurrently commute operations until operations in the same group are neighbors, and associate operations in the four groups respectively.

\begin{remark}
	To better understand Proposition~\ref{prop:space},
	let us check two examples.
	\begin{enumerate}
		\item 
		$\min(\max(\bm{h}^l,~\bm{\phi}_i^1)+\bm{\phi}_i^2-\bm{\phi}_i^3,~\bm{\phi}_i^4) \odot \bm{\phi}_i^5$
		equals to \eqref{eq:reduced} where $\hat{\bm{\phi}}_i^1=\bm{\phi}_i^1,~\hat{\bm{\phi}}_i^2=\bm{\phi}_i^4-\bm{\phi}_i^2+\bm{\phi}_i^3,~\hat{\bm{\phi}}_i^3=\bm{\phi}_i^5,~\hat{\bm{\phi}}_i^4=(\bm{\phi}_i^2-\bm{\phi}_i^3)\odot \bm{\phi}_i^5$;
		and
		
		\item 
		$\max(\min(\bm{h}^l+\bm{\phi}_i^1,~\bm{\phi}_i^2),~\bm{\phi}_i^3)\odot\bm{\phi}_i^4$ also equals to \eqref{eq:reduced} where $\hat{\bm{\phi}}_i^1=\bm{\phi}_i^3-\bm{\phi}_i^1,~\hat{\bm{\phi}}_i^2=\bm{\phi}_i^2-\bm{\phi}_i^1,~\hat{\bm{\phi}}_i^3=\bm{\phi}_i^4,~\hat{\bm{\phi}}_i^4=\bm{\phi}_i^1\odot \bm{\phi}_i^4$.
	\end{enumerate}
\end{remark}

Note that due to the
universal approximation ability of
deep network~\citep{hornik1989multilayer},
the assumption in this proposition can be easily satisfied.
For example,
we implement $A$ based on two-layer MLP in experiments (see Appendix~\ref{app:adapte}),
which can already ensure a good performance
(see Section~\ref{sec:exp-verify}).
After the transformation,
the space in \eqref{eq:reduced} also
spans a permutation-invariant binary tree, as plotted in
Figure~\ref{fig:reducesp}.
Such a space can be significantly
smaller than the original one, as 
explained in Remark~\ref{rmk:space} below.

\begin{remark}
	\label{rmk:space}

The space transformation plays an essential role in ColdNAS,
Table \ref{tab:ratio} helps to better understand to what extent 
the proposition can help reduce the search space, we take layer number $L=4$ and
the ratio is calculated as $\frac{\text{original space}}{\text{transformed space}}=\frac{6^{C\times 4}}{2^{4\times4}}$. 
\begin{table}[H]
	\centering
	\setlength\tabcolsep{2pt}
	\small
	\caption{The reduction ratio w.r.t different $C$.}
	\vspace{-5px}
		\begin{tabular}{c |c | c|c | c |c|c}
			\toprule
			$C$&	 1   &  2   & 3 &4&5&6 \\ \midrule
			Ratio&$2.0\!\times\! 10^{-2}$&$2.6\!\times\! 10^{1}$&$3.3\!\times\! 10^{4}$&$4.3\!\times\! 10^{7}$&$5.6\!\times\! 10^{10}$&$7.2\!\times\! 10^{13}$\\\bottomrule
		\end{tabular}
	\vspace{-10px}
	\label{tab:ratio}
\end{table}
Note that when $C = 1$,
the transformation will not lead to a reduction on the space.
However,
such a case is not meaningful
as it suffers from poor performance due to lack of 
flexibility for the modulation function (see Section~\ref{sec:difspace}).
The transformed spaces enable efficient and robust differentiable search, via reducing the number of architecture parameter from $6\times C\times L$ to $4\times L$. 
Meanwhile, the space size is transformed from $6^{C\times L}$ to $2^{4\times L}$, which is a reduction for any $C>1$.
\end{remark}


\subsubsection{Construction of the Supernet}
For each $M^l$, 
since there are $4$ operations at most and they are permutation-invariant, 
we only need to decide whether to take the operation or not by any order. 
We search by introducing differentiable parameters to weigh the operations and optimize the weights. For the $l$th layer of the predictor, we have 
\begin{align}\label{eq:superlayer}
	\bm{\hat{h}}^{l,k+1}\!=\!\alpha^{l,k+1}(\bm{\hat{h}}^{l,k}\circ_{\text{op}^{k+1}}\bm{\phi}_i^{k+1})
	\!+\!(1\!-\!\alpha^{l,k+1})\bm{\hat{h}}^{l,k},
\end{align}
where $\alpha^{l,k+1}$ is a weight to measure operation $ \circ_{\text{op}^{k+1}}$ in $M^l$, $k\in \{0,1,2,3\}$ and $\{ \circ_{\text{op}^{k+1}} \}_{ k= 0 }^{3}=\{\max,\min,\odot,+\}$. For notation simplicity,  we let  $\bm{\hat{h}}^{l,0}=\bm{h}^l, ~\bm{h}^l_{i}=\bm{\hat{h}}^{l,4}$.  
We construct the supernet by replacing \eqref{eq:base-mod} with \eqref{eq:superlayer}, 
i.e., replacing every $M^l$ in red dotted lines in Figure \ref{fig:mod_framework} with the structure shown in Figure \ref{fig:space} (c).

\subsubsection{Complete Algorithm}
The complete algorithm is summarized in Algorithm~\ref{alg:coldnas}. 
We first optimize the supernet and make selection to determine the structure, then reconstruct the model with determined structure and retrain it to inference.

\begin{algorithm}[ht]
	\caption{Training procedure of \TheName{}.}
	\label{alg:coldnas}
	\begin{algorithmic}[1] 
		\REQUIRE Learning rate $\beta$, number of operations to keep $K$.
		\STATE Construct the supernet by \eqref{eq:superlayer} 
		and randomly initialize all parameters $\bm{\Theta}=\{\{\alpha^{l,k}\}_{ k= 1,l=0 }^{4,L-1},\bm{\theta}_E,\bm{\theta}_A,\bm{\theta}_P\}$.
		\WHILE{Not converge}
		\FOR{Every $T_i \in \cT^{\text{train}}$ }
		\STATE Calculate $\bm{\Phi}_i$ by \eqref{eq:adapara}.
		\STATE Calculate $\hat{y}_{i,j}$ for every $v_j$ in $\cQ_i$ by \eqref{eq:forward}.
		\STATE Calculate loss $\mL_i$ by \eqref{eq:loss}.
		\ENDFOR
		\STATE $\mL^{\text{train}}=\frac{1}{|\cT^{\text{train}}|}\sum_{i=1}^{|\cT^{\text{train}}|}\mL_i$
		\STATE Update all parameters $\bm{\Theta}\leftarrow\bm{\Theta}-\beta\nabla_{\bm{\Theta}}\mL^{\text{train}}$.
		\ENDWHILE
		\STATE Determine the modulation structure by keeping operations corresponding to Top-$K$ $\alpha^{l,k}$ and remove the others.
		\STATE Construct the model with determined modulation structure and randomly initialize all parameters $\bm{\Theta}=\{\bm{\theta}_E,\bm{\theta}_A,\bm{\theta}_P\}$.
		\STATE Train the model in the same way as Step $2\sim10$.
		\STATE \textbf{Return:} The trained model.
	\end{algorithmic}
\end{algorithm}

Benefited from the great reduction brought by the space transformation, while conventional differentiable architecture search~\cite{liu2018darts} optimizes the supernet w.r.t. the bilevel objective with the upper-level variable $\bm{\alpha}$ and lower-level variable $\bm{\theta}$ :
\begin{align}\label{eq:bilevel}
\mathop{\min}_{\bm{\alpha}} \mL^{\text{val}}(\bm{\theta}^*(\bm{\alpha}),\bm{\alpha}),
\text{\;s.t.\;}
\bm{\theta}^*(\bm{\alpha})=\underset{\bm{\theta}}{\operatorname{argmin}}~ \mL^{\text{train}}(\bm{\theta},\bm{\alpha}),
\end{align}
where $\mL^{\text{val}}$ and $\mL^{\text{train}}$ represent the loss obtained on validation set and training set respectively,
we only need to optimize the supernet w.r.t. objective $\mL^{\text{train}}$ only, in an end-to-end manner by episodic training. 
For every task $T_i \in \cT^{\text{train}}$, we first input $\cS_i$ to the adaptation network $A$ to generate $\bm{\Phi}_i$ by \eqref{eq:adapara}, and then for every item $v_j \in \cQ_i$, we take $(u_i,v_j)$ and $\bm{\Phi}_i$ as input of the predictor $P$ and make prediction by \eqref{eq:forward}. 
Then, 
we use mean squared error (MSE) between the prediction $\hat{y}_{i,j}$ and true label $y_{i,j}$ as loss function:
\begin{align}\label{eq:loss}
	\mL_i=\frac{1}{M}\sum\nolimits_{j=1}^{M}(y_{i,j}-\hat{y}_{i,j})^2,
\end{align}
and $\mL^{\text{train}}=\sum\nolimits_{T_i \in \cT^{\text{train}}}\mL_i$. We update all parameters by gradient descent. 
Once the supernet converges, 
we determine all $M^l$s jointly by keeping the operation corresponding to the Top-$K$ largest values among the $4\times L$ values in $\{\alpha^{l,k+1}\}_{ k= 1,l=0 }^{4,L-1}$. 
We then retrain the model to obtain the final user cold-start model with searched modulation structure. 
During inference, a new set of tasks $\cT^{\text{test}}$ is given, which is disjoint from $\cT^{\text{train}}$. For $T_i \in \cT^{\text{test}}$, we take the whole $\cS_i$ and $(u_i,v_j)$ as input to the trained model to obtain prediction $\hat{y}_{i,j}$ for each item $v_j \in \cQ_i$.

\subsection{Discussion}

Existing works in space transformation can be divided into three types. One is using greedy search strategy~\cite{gao2021progressive}. 
Methods of this kind explore different groups of architectures
in the search space greedily.
Thus,
they fail to explore the full search space and 
can easily fall into bad local optimal.
Another is mapping the search space into a low-dimensional one.
For examples, 
using auto-encoder~\cite{Luo2018} 
or sparse coding~\cite{Yang2020}.
However,
these types of methods
do not consider special properties
of the search problem,
e.g.,
the graph structure of the supernet.
Thus,
the performance ranking
of architectures may suffer from distortion in the low-dimensional space.
ColdNAS belongs to the third type, which is to explore
architecture equivalence in the search space.
The basic idea is that if we can find a group of architectures 
that are equivalent with each other,
then evaluation any one of them is enough for all architectures in the same group.
This type of methods is problem-specific.
For example,
perturbation equivalence for matrices is explored in~\cite{Zhang2023},
morphism of networks is considered in~\cite{Jin2019}. 
The search space of ColdNAS is designed for modulation structures in cold-start problem, which has not been explored before. 
We theoretically prove the equivalence between the original space and the transformed space, which then significantly reduces the space size (Remark~\ref{rmk:space}). 
Other methods for general space reduction cannot achieve that.

\begin{table*}[t]
	\centering
	\setlength\tabcolsep{4pt}
	\small
	\caption{Test performance (\%) obtained on benchmark datasets.         
		The best results are highlighted in bold and the second-best in italic. For MSE and MAE, smaller value is better. For $\text{nDCG}_3$ and $\text{nDCG}_5$, larger value is better. }
    \vspace{-5px}
		\begin{tabular}{c|c|cccccc|cc}
			\toprule
			{Dataset}  &Metric &  DropoutNet&MeLU& MetaCS&MetaHIN&MAMO&TaNP&ColdNAS-Fixed&\TheName{}\\\midrule
			\multirow{6}{*}{MovieLens}
			&MSE&$100.90_{(0.70)}$&$95.02_{(0.03)}$&$95.05_{(0.04)}$&$91.89_{(0.06)}$&$90.20_{(0.22)}$&$\underline{89.11}_{(0.18)}$&$91.05_{(0.13)}$&$\textbf{87.96}_{(0.12)}$\\\cmidrule{2-10}
			&MAE&$85.71_{(0.48)}$&$77.38_{(0.25)}$&$77.42_{(0.26)}$&$75.79_{(0.27)}$&$	75.34_{(0.26)}$&$	\underline{74.78}_{(0.14)}$&$75.65_{(0.30)}$&$\textbf{74.29}_{(0.20)}$\\\cmidrule{2-10}
			&$\text{nDCG}_3$&$69.21_{(0.76)}$&$74.43_{(0.59)}$&$74.46_{(0.78)}$&$74.69_{(0.32)}$&$74.95_{(0.13)}$&$\underline{75.60}_{(0.07)}$&$75.11_{(0.09)}$&$\textbf{76.16}_{(0.03)}$\\\cmidrule{2-10}
			&$\text{nDCG}_5$&$68.43_{(0.48)}$&$73.52_{(0.41)}$&$73.45_{(0.56)}$&$73.63_{(0.22)}$&$73.84_{(0.16)}$&$\underline{74.29}_{(0.12)}$&$73.89_{(0.12)}$&$\textbf{74.74}_{(0.09)}$\\\midrule
			\multirow{6}{*}{BookCrossing}
			&MSE&$15.38_{(0.23)}$&$15.15_{(0.02)}$&$15.20_{(0.08)}$&$14.76_{(0.07)}$&$14.82_{(0.05)}$&${14.75}_{(0.05)}$&$ \underline{14.44}_{(0.16)}$&$\textbf{14.15}_{(0.08)}$\\\cmidrule{2-10}
			&MAE&$3.75_{(0.01)}$&$3.68_{(0.01)}$&$3.66_{(0.01)}$&$3.50_{(0.01)}$&$3.51_{(0.02)}$&$\underline{3.48}_{(0.01)}$&$3.49_{(0.02)}$&$\textbf{3.40}_{(0.01)}$\\\cmidrule{2-10}
			&$\text{nDCG}_3$&$77.66_{(0.18)}$&$\underline{77.69}_{(0.15)}$&$77.68_{(0.12)}$&$77.66_{(0.19)}$&$	77.68_{(0.09)}$&$77.48_{(0.06)}$&$77.65_{(0.09)}$&$\textbf{77.83}_{(0.01)}$\\\cmidrule{2-10}
			&$\text{nDCG}_5$&$80.87_{(0.15)}$&$	81.10_{(0.15)}$&$80.97_{(0.09)}$&$	80.95_{(0.04)}$&$	81.01_{(0.05)}$&$\underline{81.16}_{(0.21)}$&$81.12_{(0.06)}$&$\textbf{81.32}_{(0.10)}$\\\midrule
			\multirow{6}{*}{Last.fm}
			&MSE&$21.91_{(0.38)}$&$21.69_{(0.34)}$&$21.68_{(0.12)}$&$\underline{21.43}_{(0.23)}$&$21.64_{(0.10)}$&$ 21.58_{(0.20)}$&$21.62_{(0.16)}$&$ \textbf{20.91}_{(0.05)}$\\\cmidrule{2-10}
			&MAE&$43.02_{(0.52)}$&$	42.28_{(1.21)}$&$42.28_{(0.76)}$&$\underline{42.07}_{(0.49)}$&$42.30_{(0.28)}$&$42.15_{(0.56)}$&$42.32_{(0.34)}$&$\textbf{41.78}_{(0.24)}$\\\cmidrule{2-10}
			&$\text{nDCG}_3$&$75.13_{(0.48)}$&$80.15_{(2.09)}$&$80.81_{(0.97)}$&$\underline{82.01}_{(0.56)}$&$80.73_{(0.80)}$&$81.03_{(0.36)}$&$80.77_{(0.32)}$&$\textbf{82.80}_{(0.69)}$\\\cmidrule{2-10}
			&$\text{nDCG}_5$&$69.03_{(0.31)}$&$75.03_{(0.68)}$&$75.01_{(0.64)}$&$\underline{75.98}_{(0.33)}$&$75.45_{(0.29)}$&$\underline{75.98}_{(0.41)}$&$75.48_{(0.21)}$&$\textbf{76.77}_{(0.10)}$
			\\\bottomrule
		\end{tabular}
	\label{tab:results}
\end{table*}

\begin{table*}[ht]
	\centering
	\setlength\tabcolsep{3pt}
	\small
	\caption{Modulation structure with Top-$4$ operations searched on the three benchmark datasets respectively. We also show the modulation structure of ColdNAS-Fixed, which is the same regardless of the dataset used.}
	\vspace{-5px}
		\begin{tabular}{c |c | c|c | c }
			\toprule
			&	 $M^0$    & $M^1$   &  $M^2$ &$M^3$\\ \midrule
			{MovieLens}&$\min(\max(\bm{h}^0,\bm{\phi}_i^{0,1}),\bm{\phi}_i^{0,2})+\bm{\phi}_i^{0,3}$ &$\bm{h}^1+\bm{\phi}_i^{1,1}$&$\bm{h}^2$&$\bm{h}^3$\\ \midrule
			{BookCrossing}&$\min(\bm{h}^0,\bm{\phi}_i^{0,1})$&$\bm{h}^1+\bm{\phi}_i^{1,1}$&$\bm{h}^2\odot\bm{\phi}_i^{2,1}+\bm{\phi}_i^{2,2}$&$\bm{h}^3$\\ \midrule
			{Last.fm}&$\bm{h}^0+\bm{\phi}_i^{0,1}$&$\bm{h}^1+\bm{\phi}_i^{1,1}$&$\max(\bm{h}^2,\bm{\phi}_i^{2,1})+\bm{\phi}_i^{2,2}$&$\bm{h}^3$\\\midrule
			{ColdNAS-Fixed}&$\bm{h}^0\odot\bm{\phi}_i^{0,1}+\bm{\phi}_i^{0,2}$&$\bm{h}^1\odot\bm{\phi}_i^{1,1}+\bm{\phi}_i^{1,2}$&$\bm{h}^2\odot\bm{\phi}_i^{2,1}+\bm{\phi}_i^{2,2}$&$\bm{h}^3\odot\bm{\phi}_i^{3,1}+\bm{\phi}_i^{3,2}$\\ \bottomrule
		\end{tabular}
	\label{tab:searched_structure}
\end{table*}

\section{Experiments}

We perform experiments on three benchmark datasets with 
the aim to answer the following research questions:
\begin{itemize}[leftmargin=*]
	\item \textbf{RQ1}: What is the modulation structure selected by ColdNAS and how does \TheName{} perform in comparison with the state-of-the-art cold-start models?
	\item \textbf{RQ2}: How can we understand the search space and algorithm of ColdNAS?
	\item \textbf{RQ3}: How do hyperparameters affect \TheName{}?
\end{itemize}
Results 
are averaged over five runs.

%
%
%
%
%
%

\subsection{Datasets}

We use three benchmark datasets (Table~\ref{tab:data-stat}):  
(i) \textbf{MovieLens}~\cite{harper2015movielens}: a dataset containing 1 million movie ratings of users collected from MovieLens, whose features include gender, age, occupation, Zip code, publication year, rate, genre, director and actor; 
(ii) \textbf{BookCrossing}~\cite{ziegler2005improving}: a collection of users' ratings on books in BookCrossing community, whose features include age, location, publish year, author, and publisher; 
and 
(iii) \textbf{Last.fm}: a collection of user's listening count of artists from Last.fm online system, whose features only consist of user and item IDs. 
Following \citet{lin2021task}, we generate negative samples for the query sets in Last.fm. 
\begin{table}[ht]
	\centering
	\setlength\tabcolsep{2pt}
		\small
		\caption{Summary of datasets used in this paper.}
		\vspace{-5px}
		\begin{tabular}{c |c | c|c|c|c}
			\toprule
			Dataset&	 \# User (Cold)    &  \# Item   &  \# Rating&\# User Feat.&\# Item Feat. \\ \midrule
			{MovieLens}&6040 (52.3\%)&3706&1000209&4&5\\ \midrule
			{BookCrossing}&278858 (18.6\%)&271379&1149780&2&3\\ \midrule
			{Last.fm}&1872 (15.3\%)&3846&42346&1&1\\\bottomrule
		\end{tabular}
	\label{tab:data-stat}
	\vspace{-5px}
\end{table}

\subsubsection*{Data Split} 
Following \citet{lin2021task}, the ratio of
$\cT^{\text{train}}:\cT^{\text{val}}:\cT^{\text{test}}$ is set as $7:1:2$.  $\cT^{\text{val}}$ is used to judge the convergence of supernet. $\cT^{\text{train}},\cT^{\text{val}},\cT^{\text{test}}$ contain no overlapping users. 
For  MovieLens and Last.fm, we keep any user whose interaction history length lies in $[40,200]$. 
Each support set contains $N=20$ randomly selected interactions of a user, and query set contains the rest  interactions of the same user. 
As for BookCrossing with severe long-tail distribution of user-item interactions, 
we particularly put any user  whose interaction history length lies in $[50,1000)$ into $\cT^{\text{train}}$.  Then, we divide users with interaction history length in $[2,50)$ into $70\%$, $10\%$ and $20\%$ to be put in  $\cT^{\text{train}}$, $\cT^{\text{val}}$, $\cT^{\text{test}}$ respectively. The proportion of cold users in each dataset is also shown in Table~\ref{tab:data-stat}.
Then, we randomly select half of each user's interaction history as support set and take the rest as query set. 

\subsubsection*{Evaluation Metric} 
Following~\cite{lee2019melu,pang2022pnmta}, 
we evaluate the performance by 
mean average error (MAE), mean squared Error (MSE), 
normalized discounted cumulative gain $\text{nDCG}_3$ and $\text{nDCG}_5$. MAE and MSE evaluate the numerical gap between the prediction and the ground-truth rating, lower value is better. For $\text{nDCG}_3$ and $\text{nDCG}_5$, the higher value is better, representing the proportion between the discounted cumulative gain of the predicted item list and the ground-truth list.

\subsection{Performance Comparison (RQ1)} 
\label{sec:exp:perf}

\begin{figure*}[ht]
	\vspace{-10px}
	\centering
	\subfigure[MovieLens.]
	{\includegraphics[width=0.28\textwidth]{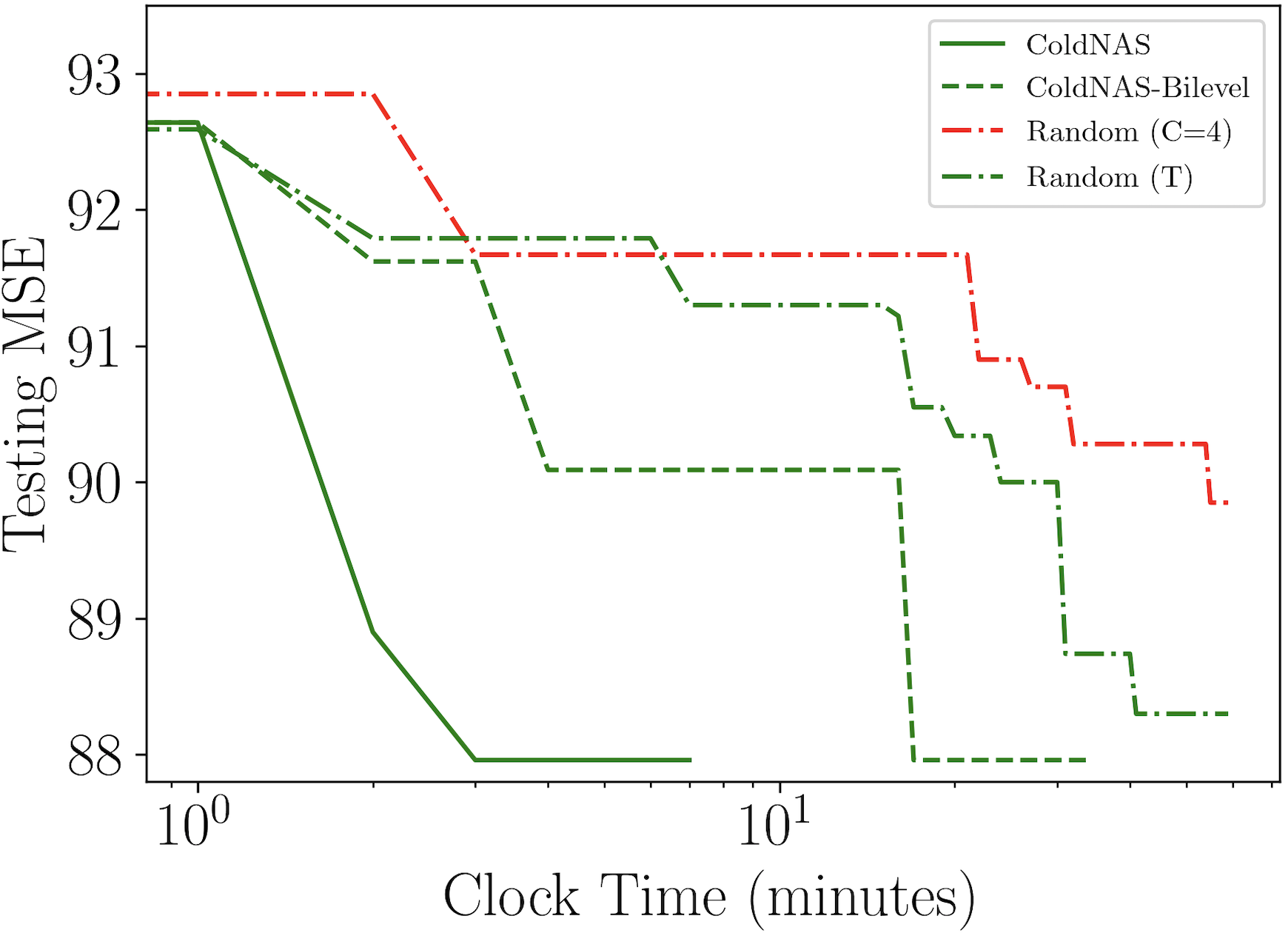}
		\label{fig:mls}}
	\quad
	\subfigure[BookCrossing.]
	{\includegraphics[width=0.28\textwidth]{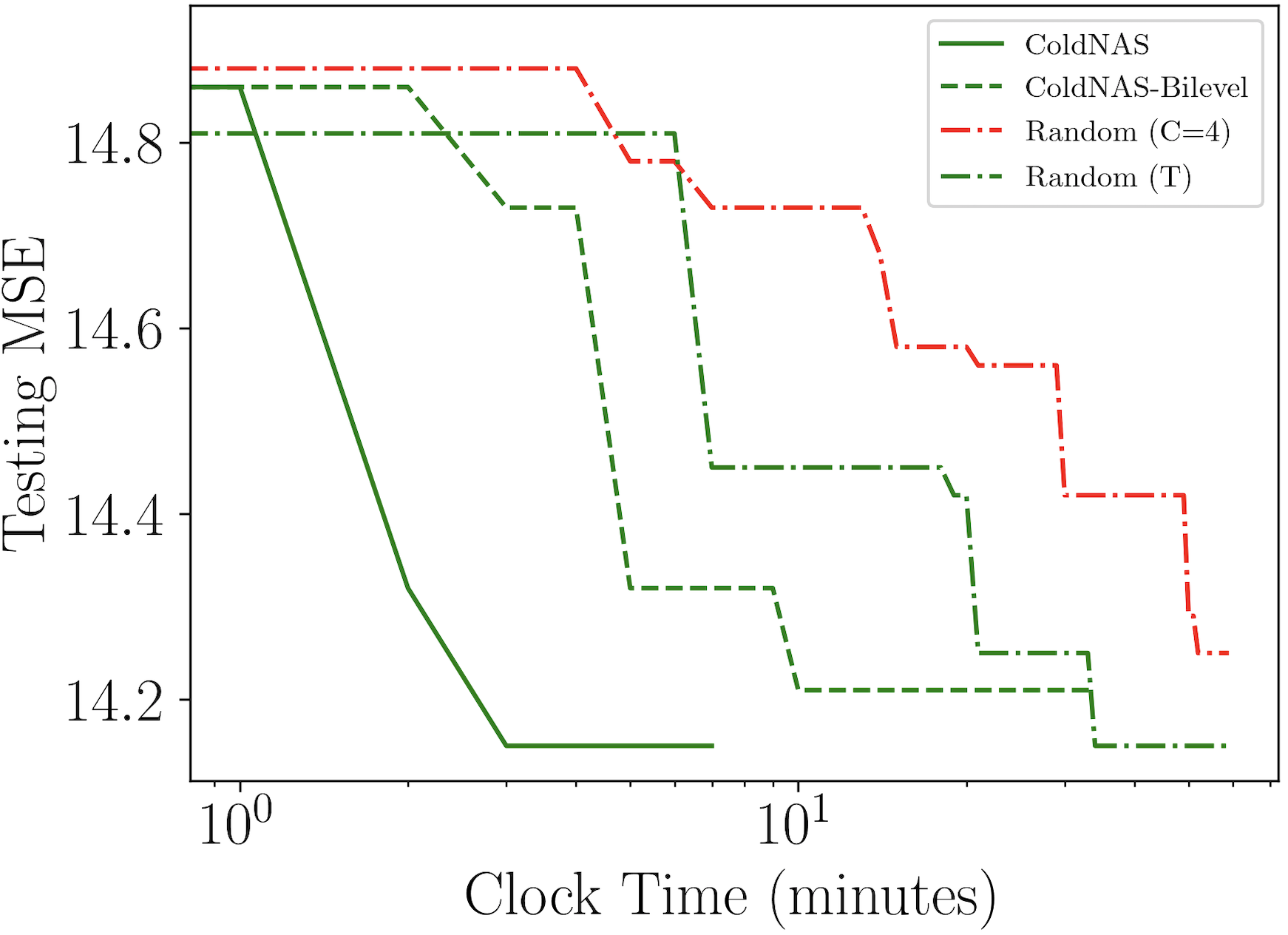}
		\label{fig:bxs}}
	\quad
	\subfigure[Last.fm.]
	{\includegraphics[width=0.28\textwidth]{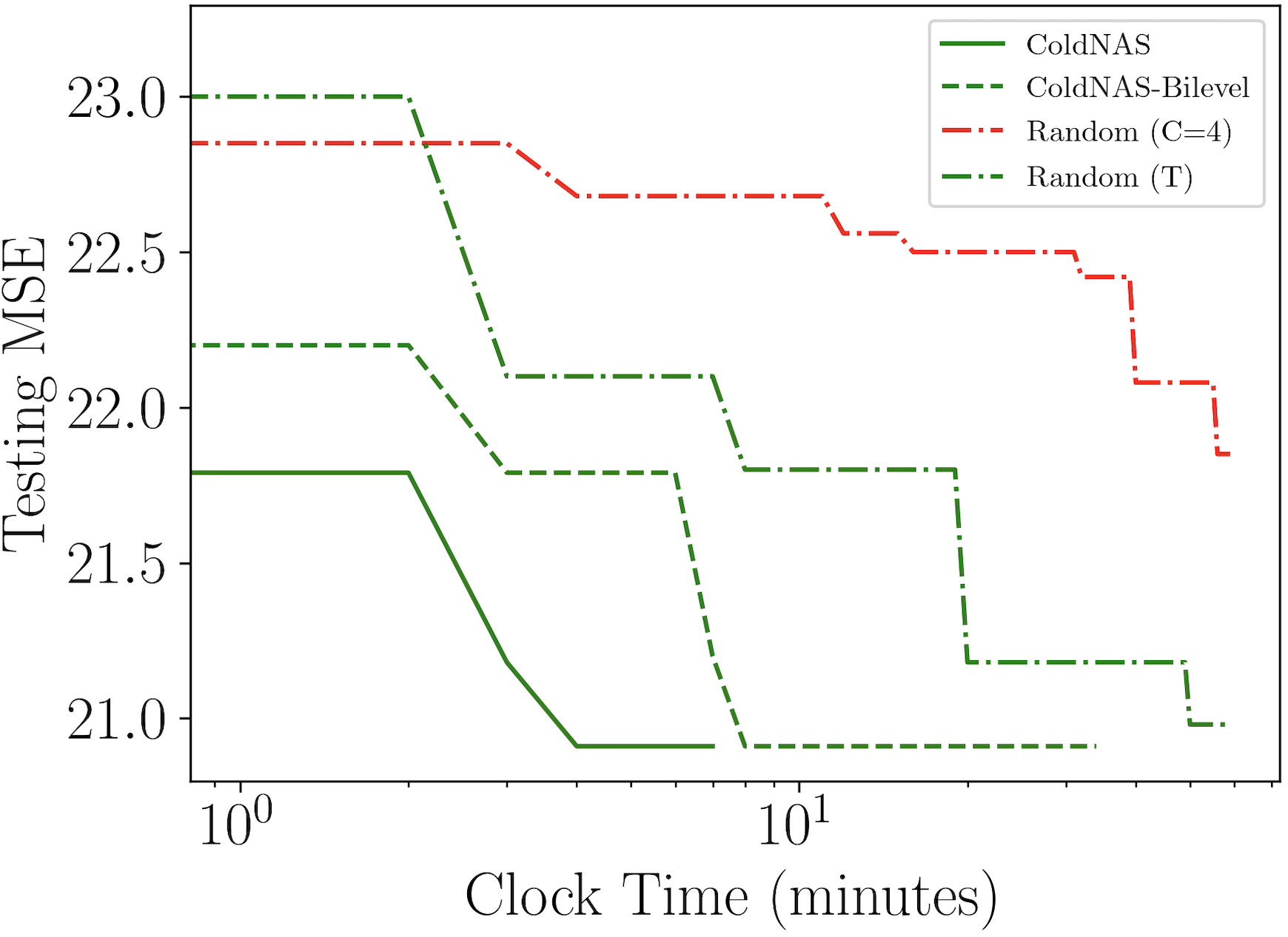}
		\label{fig:lfms}}
	\vspace{-13px}
	\caption{Testing MSE vs clock time of different search strategies. ColdNAS, ColdNAS-bilevel and Random-t operate on the transformed space, while Random operates on the original space with $C=4$ in every $M^l$.}
	\label{fig:search}
	\vspace{-10px}
\end{figure*}

We compare \textbf{\TheName{}} with the following representative user cold-start methods: 
(i)
traditional deep cold-start model 
\textbf{DropoutNet}~\cite{volkovs2017dropoutnet} 
and
(ii) FSL based methods include
\textbf{MeLU}~\cite{lee2019melu},
\textbf{MetaCS} \cite{bharadhwaj2019meta}, 
\textbf{MetaHIN}~\cite{lu2020meta}, 
\textbf{MAMO}~\cite{dong2020mamo}, 
and 
\textbf{TaNP}~\cite{lin2021task}.  
We run the public codes provided by the respective authors. 
PNMTA~\cite{pang2022pnmta}, 
CMML~\cite{feng2021cmml}, 
PAML~\cite{wang2021preference} and 
REG-PAML~\cite{yu2021personalized} are not compared due to the lack of public codes. We choose a 4-layer predictor, more details of our model and parameter setting are provided in Appendix~\ref{app:expt-setting}.
We also compare with a variant of ColdNAS called 
\textbf{ColdNAS-Fixed},
which uses the fixed FiLM function in \eqref{eq:film} at every layer rather than our searched modulation function.


Table~\ref{tab:results} shows the overall user-cold start recommendation performance for all methods.
We can see that
ColdNAS significantly outperforms the others on all the datasets and metrics. 
Among all compared baselines, 
DropoutNet performs the worst as it is not a few-shot learning method that the model has no ability to adapt to different users. 
Among meta-learning based methods, 
MeLU, MetaCS, MetaHIN and MAMO adopt gradient-based meta-learning strategy, which may suffer from 
overfitting during local-updates. 
In contrast, TaNP and \TheName{} learn to generate user-specific parameters to guide the adaptation. 
TaNP uses a fixed modulation structure which may not be optimal for different datasets, while 
ColdNAS automatically finds the optimal structure. 
Further,  the consistent performance gain of ColdNAS over ColdNAS-Fixed validates the necessity of searching modulation structure to fit datasets instead of using a fixed one. 

Table \ref{tab:searched_structure} shows 
the searched modulation structures.
We find that the searched modulation structures are different across the datasets. 
No modulation should be taken at the last layer, as directly use user-specific preference possibly leads to overfitting.
Note that all the operations in the transformed search space have been selected at least once, 
which means all of them are helpful.

\begin{figure*}[ht]
	\centering
	\subfigure[MovieLens.]
	{\includegraphics[width=0.28\textwidth]{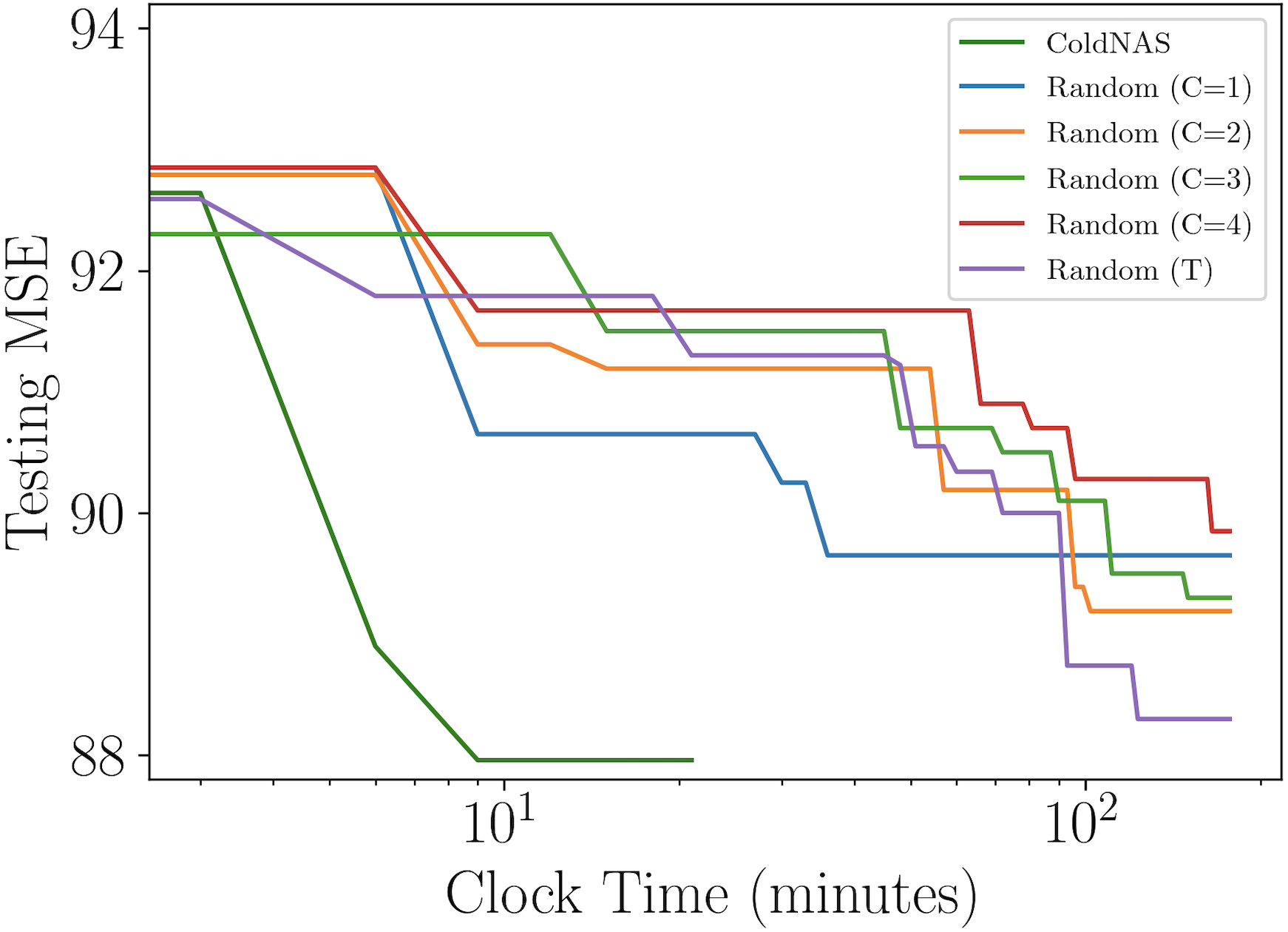}
		\label{fig:mlc}}
	\quad
	\subfigure[BookCrossing.]
	{\includegraphics[width=0.28\textwidth]{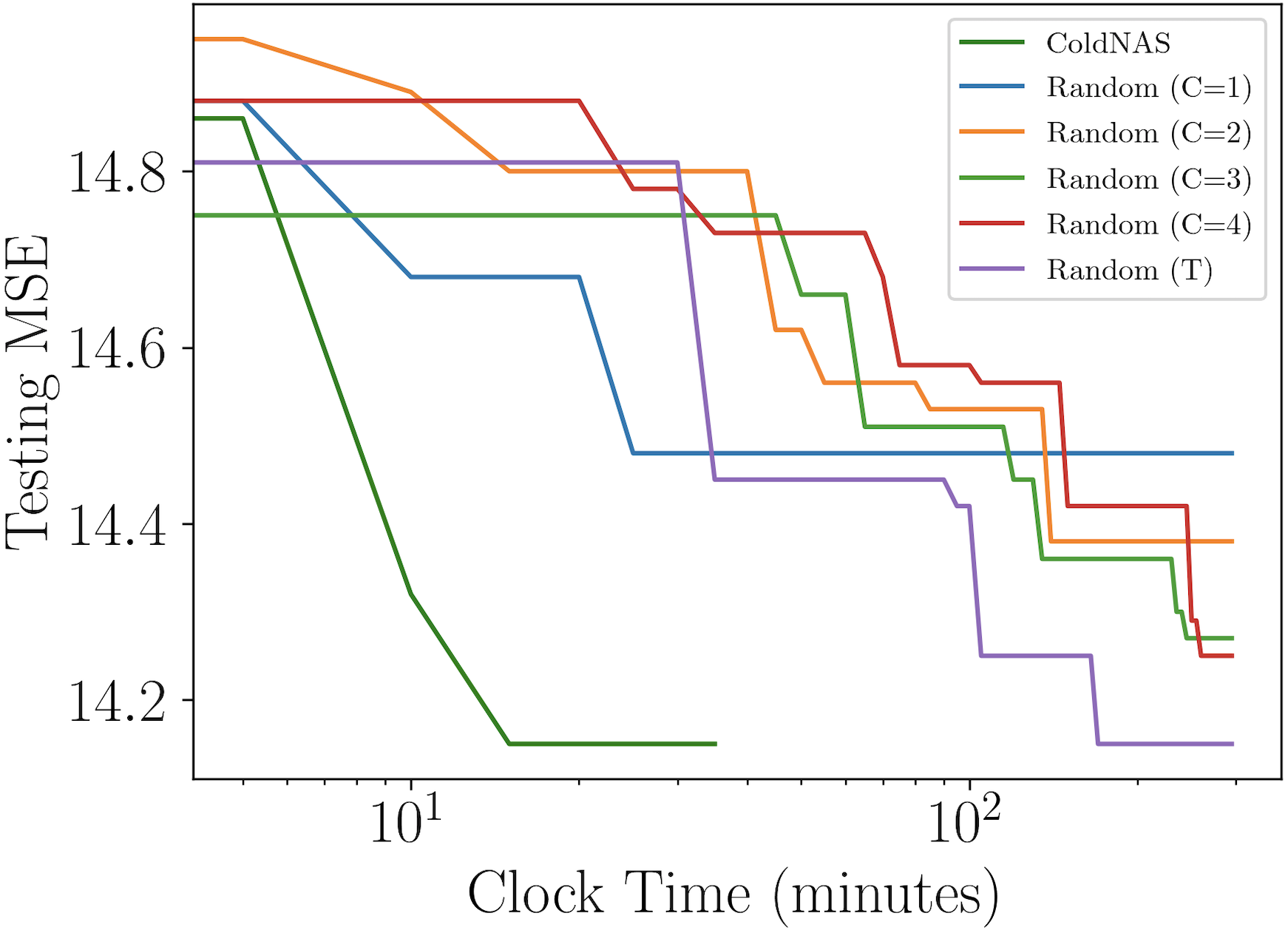}
		\label{fig:bxc}}
	\quad
	\subfigure[Last.fm.]
	{\includegraphics[width=0.28\textwidth]{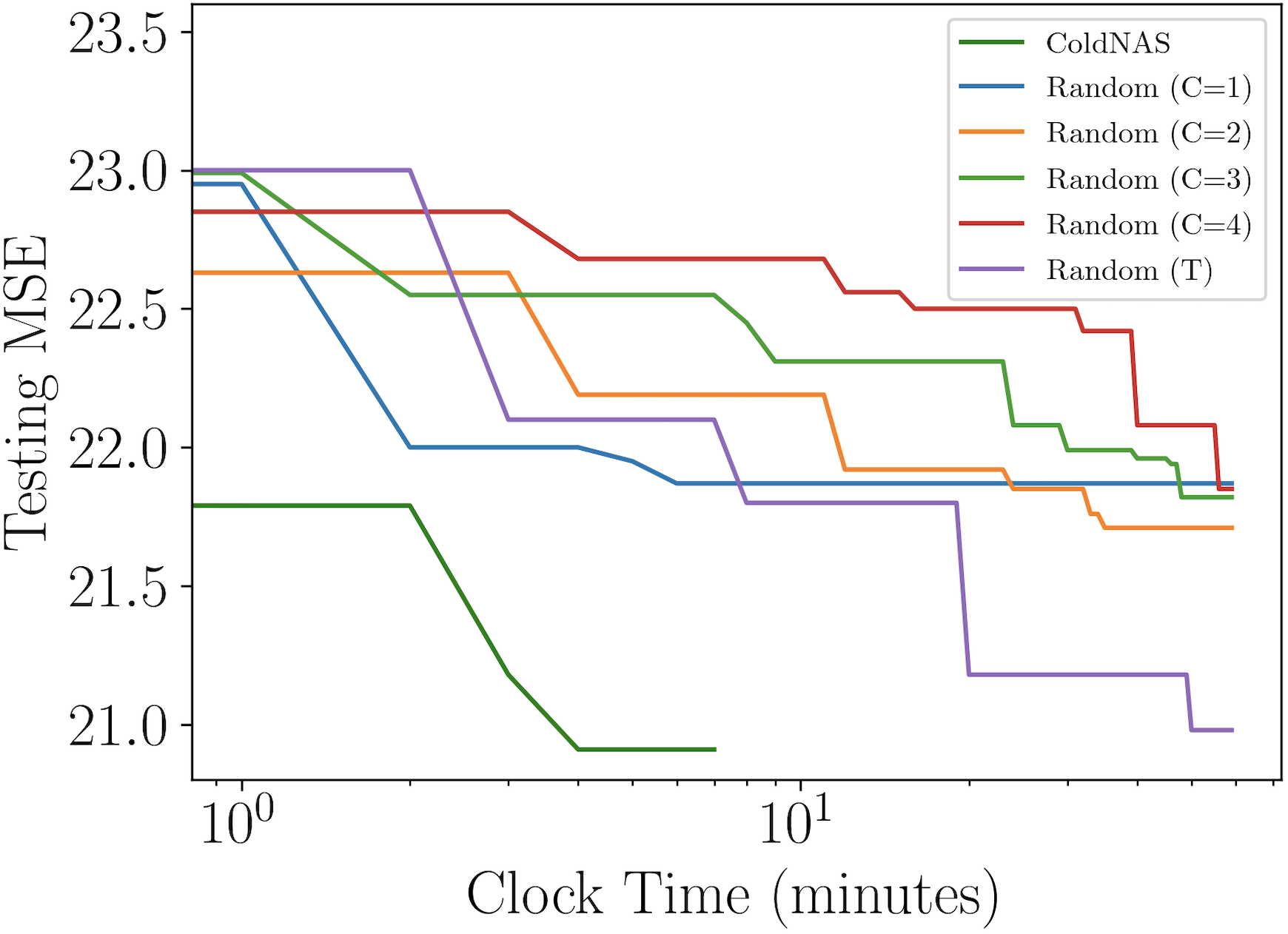}
		\label{fig:lfmc}}
	\vspace{-10px}
	\caption{Testing MSE vs clock time on various search spaces: transformed space, and original spaces with different $C$.}
	\label{fig:c}
\end{figure*}

Comparing time consumption to other cold-start models, 
ColdNAS only additionally optimizes a supernet with the same objective and comparable size. 
Table \ref{tab:time} shows 
the clock time taken by \TheName{} and TaNP which obtains the second-best in Table~\ref{tab:results}. 
As can be observed, 
the searching in \TheName{} is very efficient, as it is only slightly more expensive
than retraining. 

\begin{table}[htbp]
	\vspace{-5px}
	\centering
	\setlength\tabcolsep{3pt}
	\small
	\caption{Clock time taken by \TheName{} and TaNP.}
	\vspace{-5px}
		\begin{tabular}{c|c |c | c|c  }
			\toprule
			\multicolumn{2}{c|}{Clock time (min)}   & MovieLens   &  BookCrossing &Last.fm\\ \midrule
			\multicolumn{2}{c|}{TaNP}&15.5&44.2&4.1\\ \midrule
			\multirow{2}{*}{ColdNAS}&Search&16.2&45.5&3.9\\ \cmidrule{2-5}
			&Retrain&12.7&35.5&3.5\\ \bottomrule
		\end{tabular}
	\label{tab:time}
	\vspace{-10px}
\end{table}

\subsection{Searching in \TheName{} (RQ2)}

\subsubsection{Choice of Search Strategy}
In \TheName{}, we optimize \eqref{eq:loss} by gradient descent. 
In this section, we compare \TheName{} with other search strategies to search the modulation structure, including \textbf{Random ($\bm{C=4}$)} which conducts random search \cite{bergstra2012random} on the original space with 
operation number $C=4$ in each $M^l$, 
\textbf{Random (T)} which conducts random search \cite{bergstra2012random} on the transformed space, \textbf{ColdNAS-Bilevel} which optimizes the supernet with the bilevel objective \ref{eq:bilevel}.
For random search, we record its training and evaluation time. 
For ColdNAS and ColdNAS-bilevel, we sample the architecture parameter of the supernet and retrain from scratch, and exclude the time for retraining and evaluation. 
Figure \ref{fig:search} shows the performance of the best searched architecture. 
One can observe that searching on the transformed space allows higher efficiency and accuracy, comparing Random (T) to Random ($C=4$). 
In addition, differentiable search on the supernet structure has brought more efficiency, as both ColdNAS and ColdNAS-Bilevel outperform random search. 
ColdNAS and ColdNAS-Bilevel converge to similar testing MSE, but ColdNAS is faster. 
This validates the efficacy of directly using gradient descent on all parameters in ColdNAS. 
\begin{figure*}[t]
	\vspace{-5px}
	\centering
	\subfigure[Varying $K$ in Top-$K$.\label{fig:last-k}]{
		\includegraphics[width=0.28\textwidth]{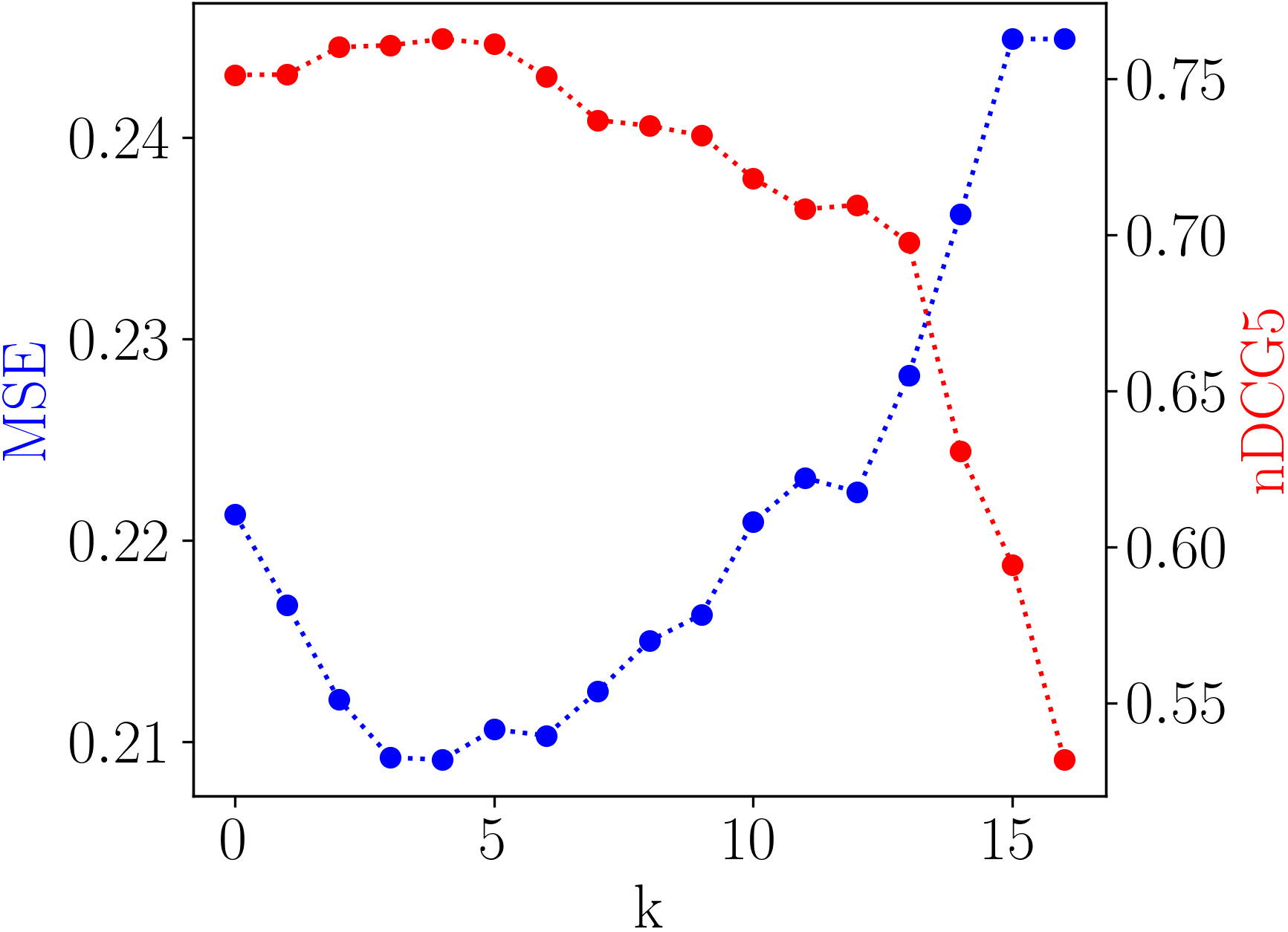}}
	\quad
	\subfigure[Varying number of layers of the predictor.\label{fig:last-mlp}]{
		\includegraphics[width=0.28\textwidth]{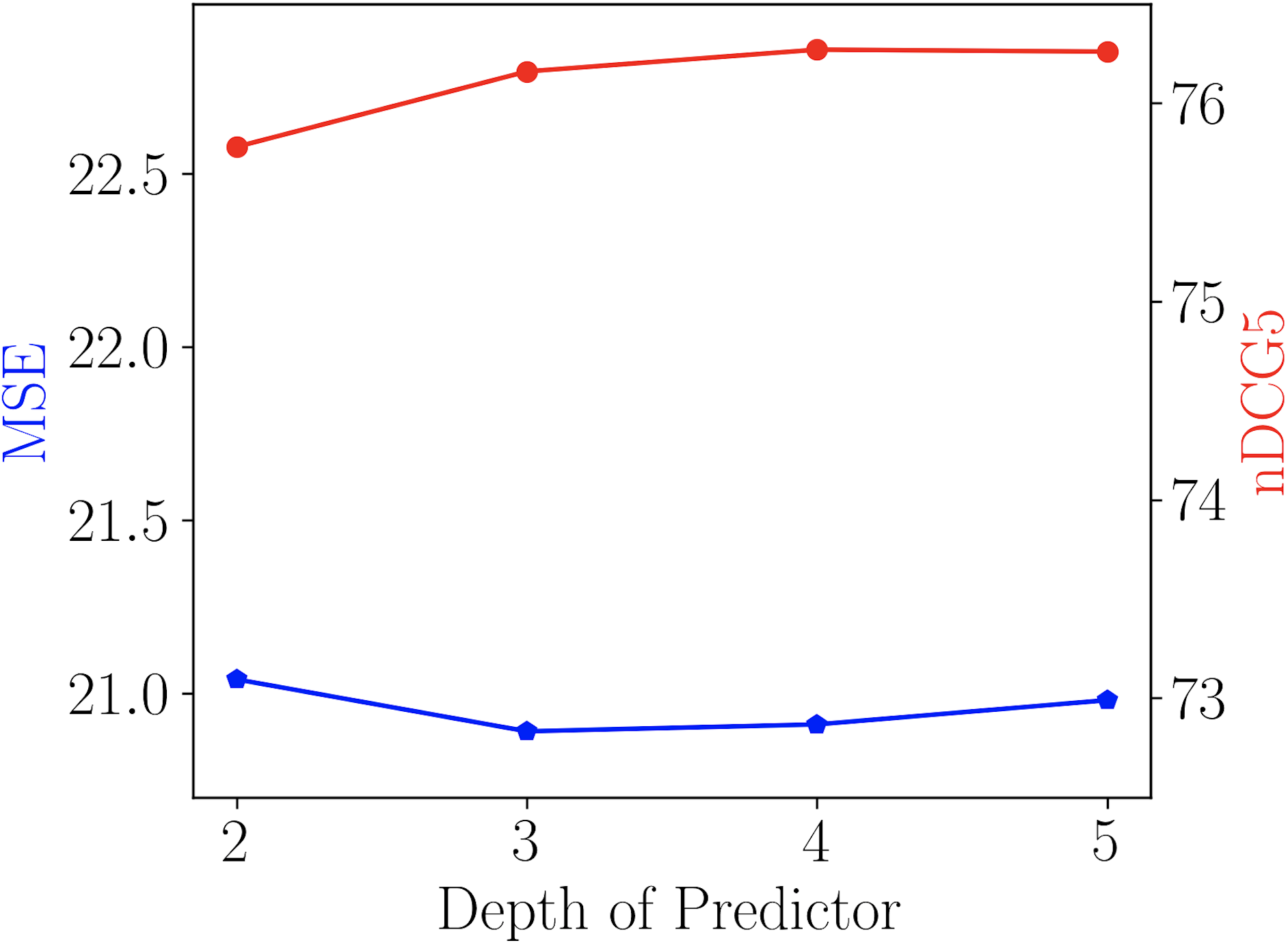}}
	\quad
	\subfigure[Varying support set size during inference.\label{fig:last-support}]{
		\includegraphics[width=0.28\textwidth]{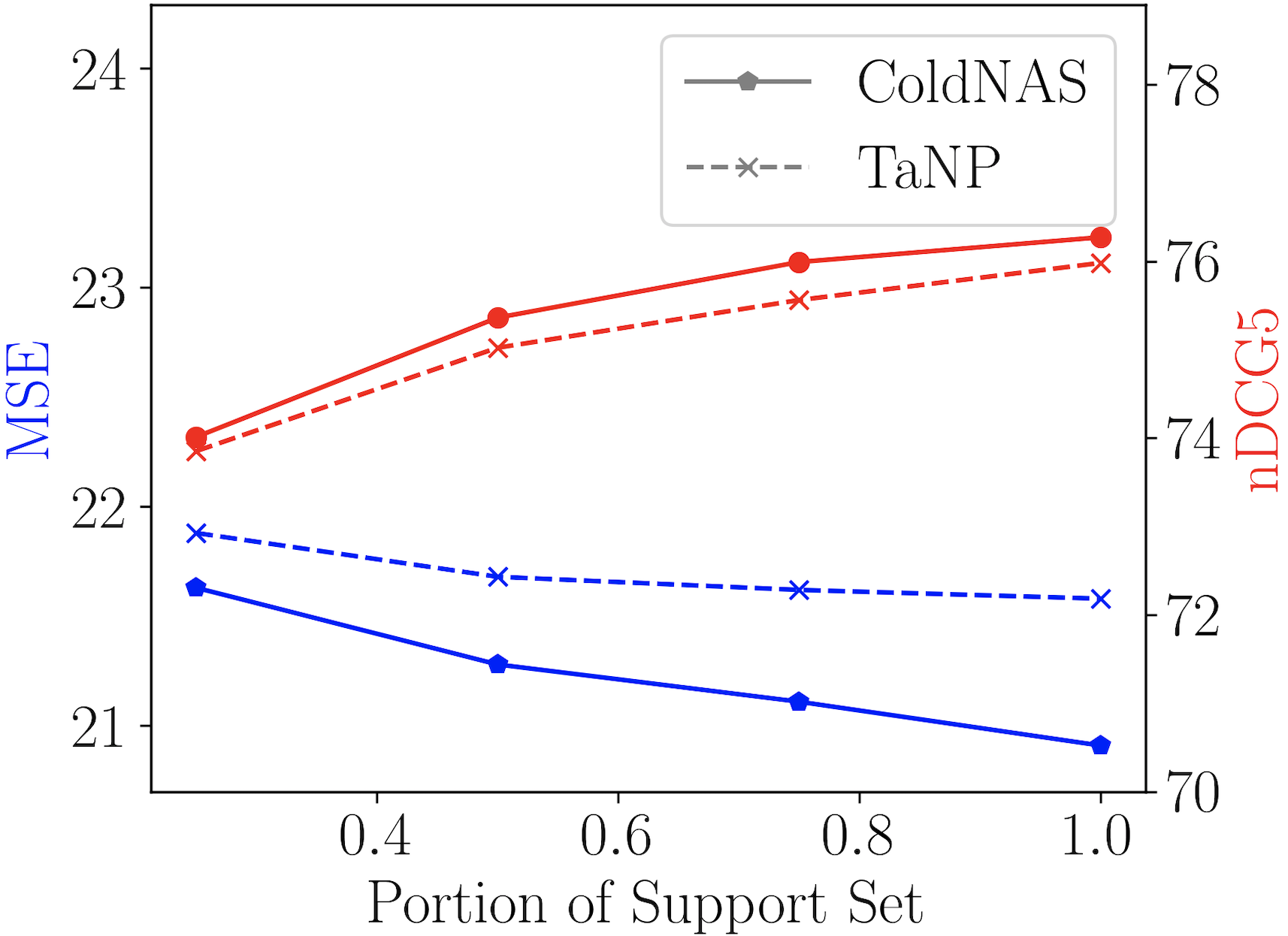}}
	\vspace{-10px}
	\caption{Model sensitivity analysis of \TheName{} on Last.fm.}
	\label{fig:sensitivity-last}
\end{figure*}
\subsubsection{Necessity of Search Space Transformation}
\label{sec:difspace}
Now that the search algorithm has been chosen, we now particularly examine the necessity of search space transformation in terms of time and performance. 
As shown in Table \ref{tab:ratio}, a larger $C$ which constrains the number of operations will lead to a larger original space. 
And the space reduction ratio grows exponentially along with the original space size. 
Figure \ref{fig:c} plots the testing MSE vs clock time of
random search on original spaces of different size (Random ($C=c$)), random search on transformed space (Random (T)), 
and our ColdNAS. 
When $C$ is small, modulation functions would not be flexible enough,  though it is
easy to find the optimal candidate in this small search space. 
When $C$ is large, the search space is large, where 
random search would be too time-consuming to find a good modulation structure. 
In contrast, 
the transformed search space has a consistent small size, and is theoretically proved to be equal to the space with any $C$. As can be seen, both Random (T)
and ColdNAS can find good structure more effective than Random ($C=4$), and
ColdNAS searches the fastest via differentiable search.

\subsubsection{Understanding Proposition~\ref{prop:space}}
\label{sec:exp-verify}
\begin{figure}[ht]
	\vspace{-10px}
	\centering
	
	\subfigure[2 layers (chosen).]{
		\includegraphics[width=0.14\textwidth]{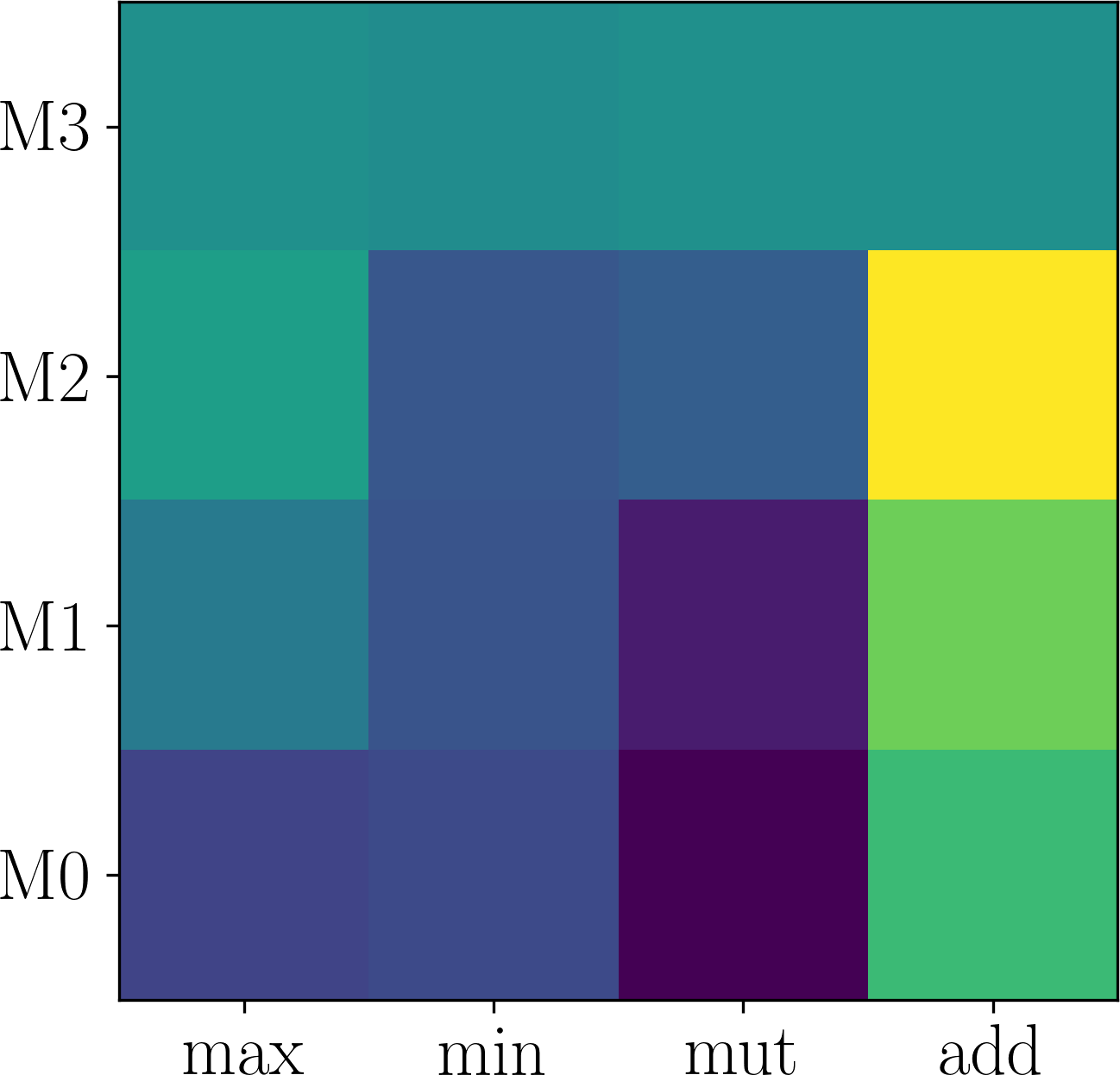}}
	\subfigure[3 layers.]{
		\includegraphics[width=0.14\textwidth]{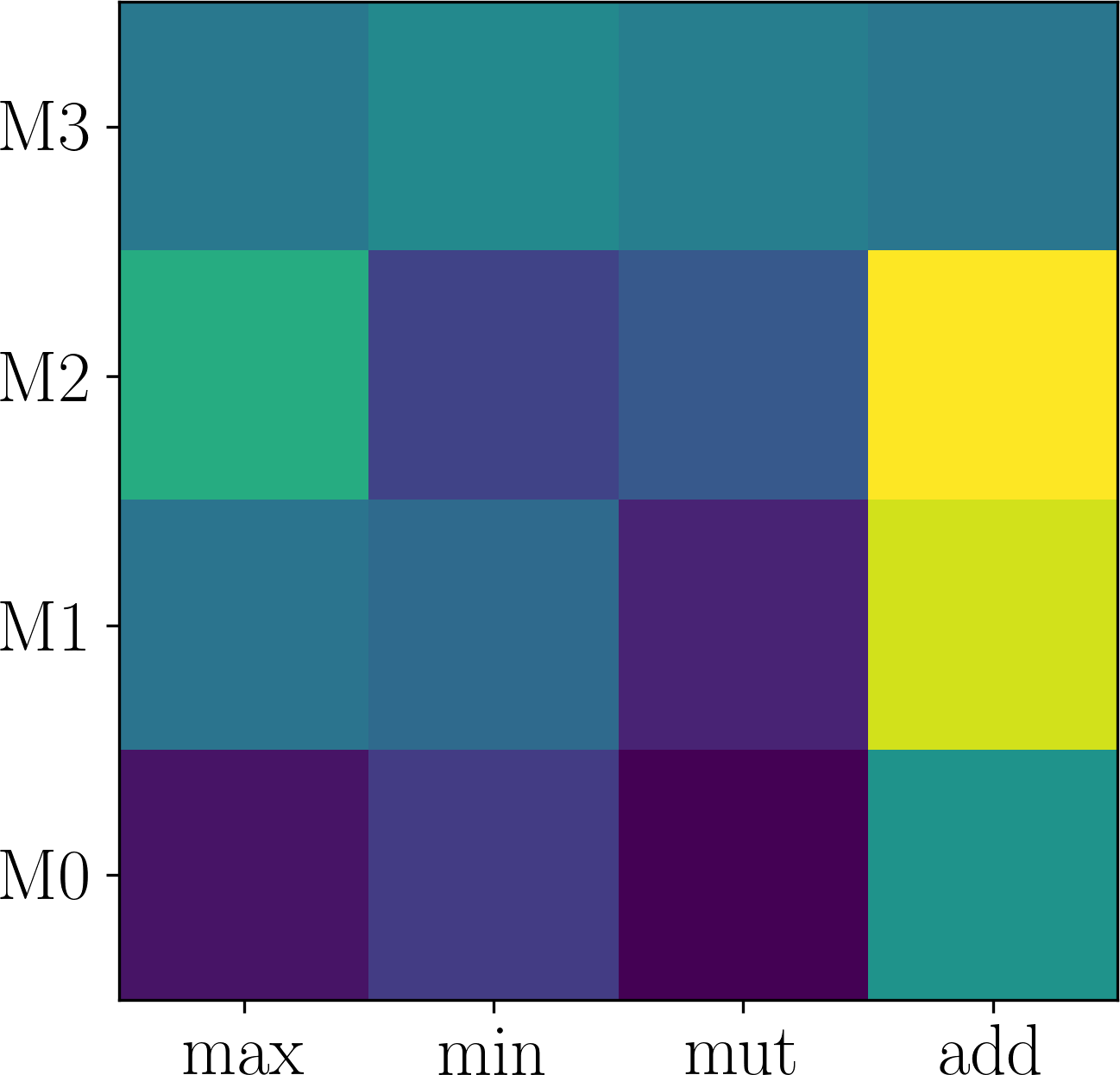}}
	\subfigure[4 layers.]{
		\includegraphics[width=0.14\textwidth]{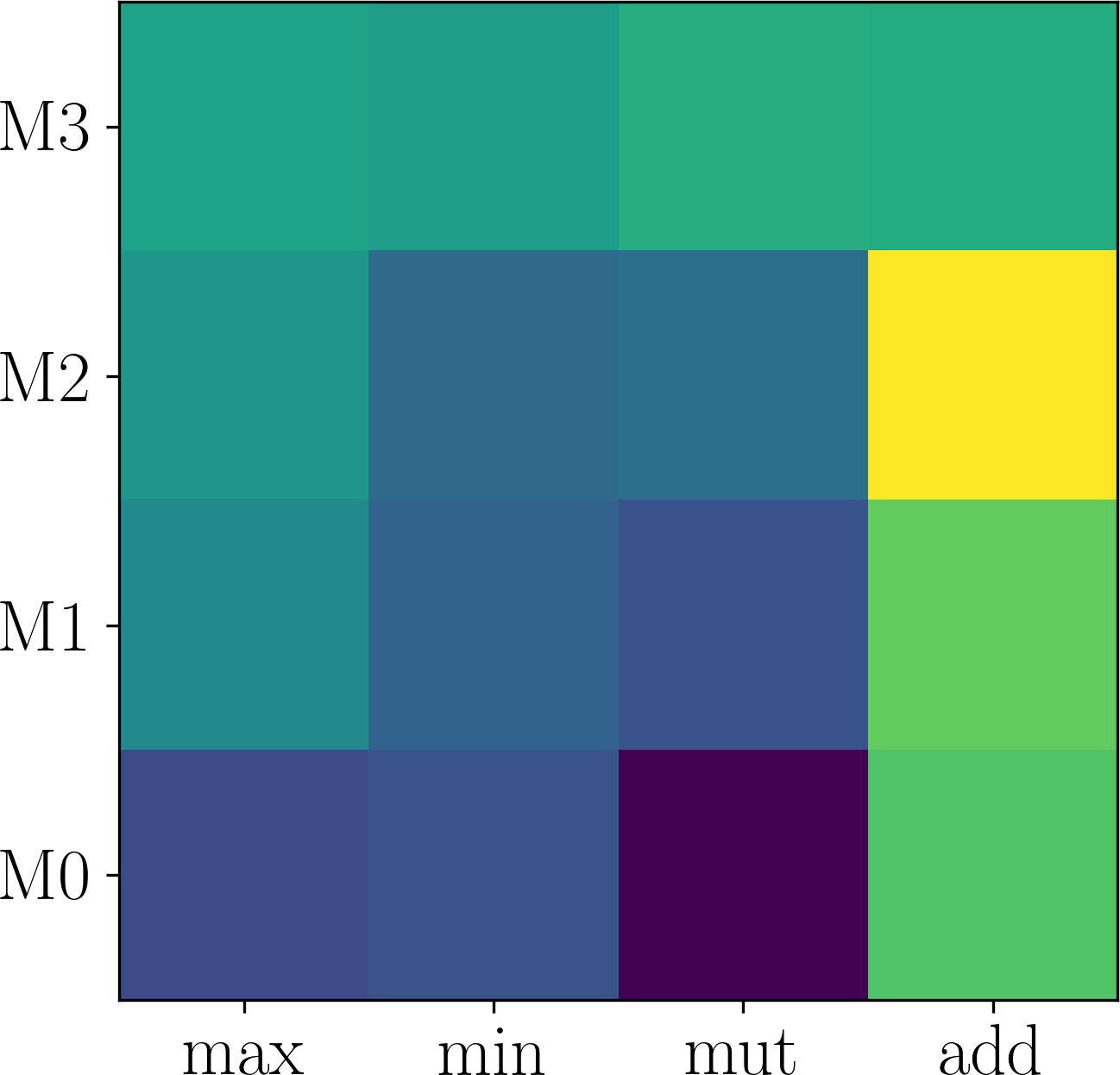}}
	
	\subfigure[5 layers.]{
		\includegraphics[width=0.164\textwidth]{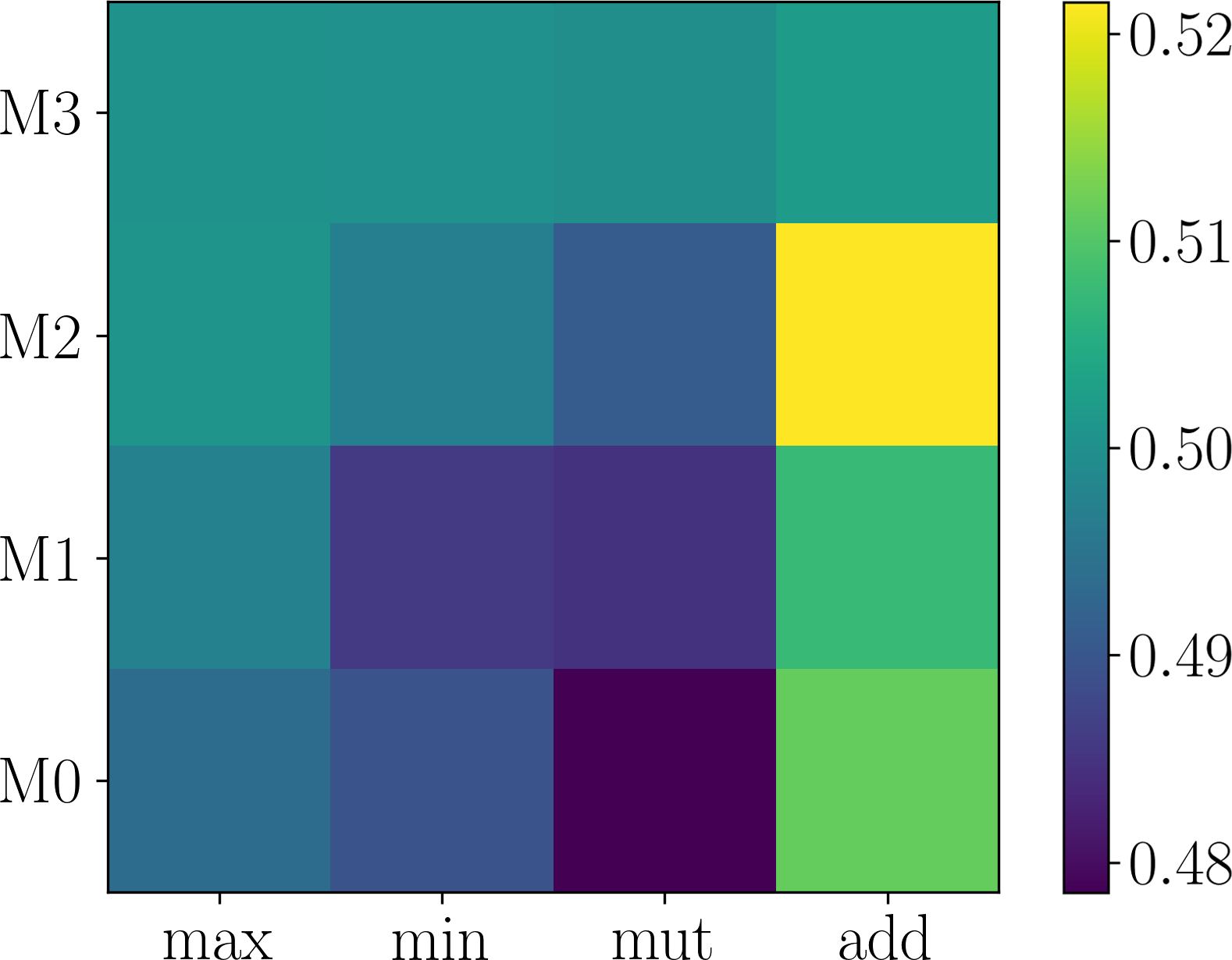}}
	\subfigure[Performance.]{
		\includegraphics[width=0.270\textwidth]{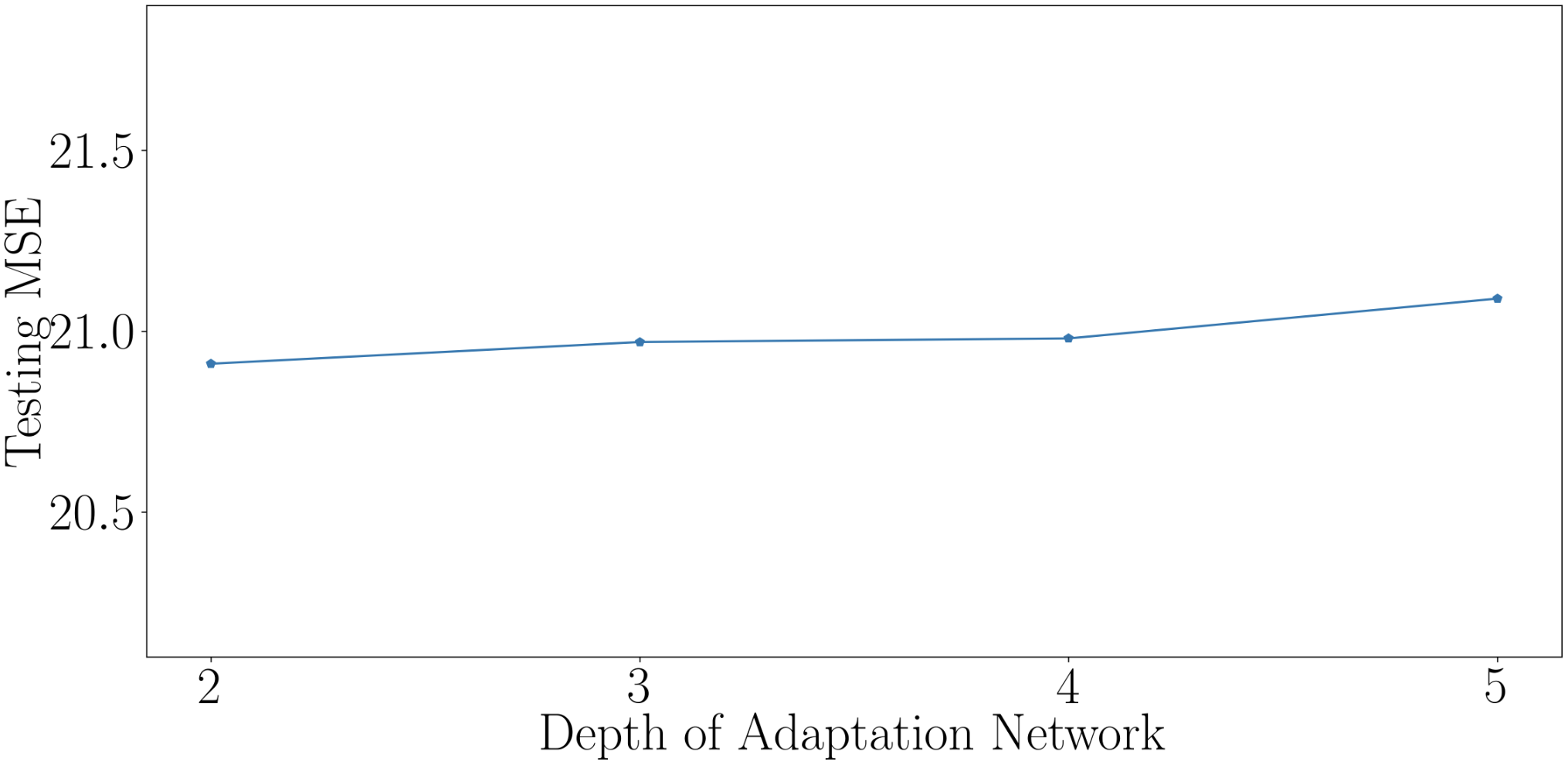}}
	\vspace{-10px}
	\caption{Effect of changing the depth of adaptation network on Last.fm. 
		Figure~\ref{fig:prop-assump} (a-d) shows the visualization of the $\{\alpha^{l,k}\}$ obtained with varying depth, and Figure~\ref{fig:prop-assump} (e) shows the testing MSE obtained correspondingly. 
}
	\label{fig:prop-assump}
	\vspace{-10px}
\end{figure}

Here, we first show that the assumption of adaptation network $A$ being expressive enough can be easily satisfied. 
Generally,
deeper networks are more expressive.
Figure~\ref{fig:prop-assump} plots the effect of changing the depth of the neural network in $A$ 
on Last.fm.  
As shown in Figure \ref{fig:prop-assump} (a-d), although different number of layers in $A$ are used, the searched modulation operations are the same. 
Consequently, the performance difference is small, as shown in Figure~\ref{fig:prop-assump} (d). 
Thus, we use $2$ layers which is expressive enough and has smaller parameter size. 

\begin{figure}[ht]
	\vspace{-10px}
	\centering
	\subfigure[Changing + to -.]{
		\includegraphics[width=0.14\textwidth]{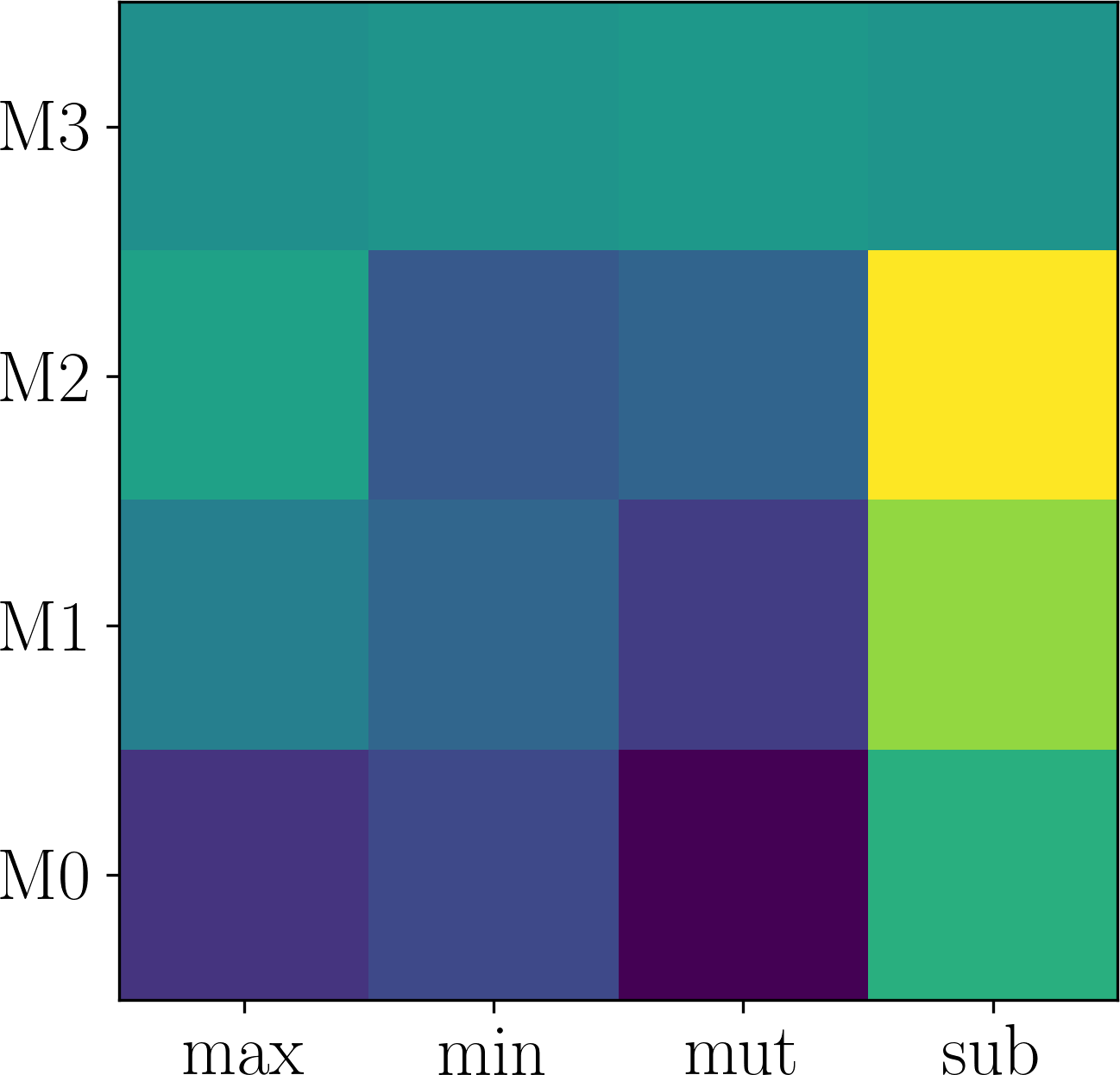}}
	\subfigure[Changing $\odot$ to /.]{
		\includegraphics[width=0.14\textwidth]{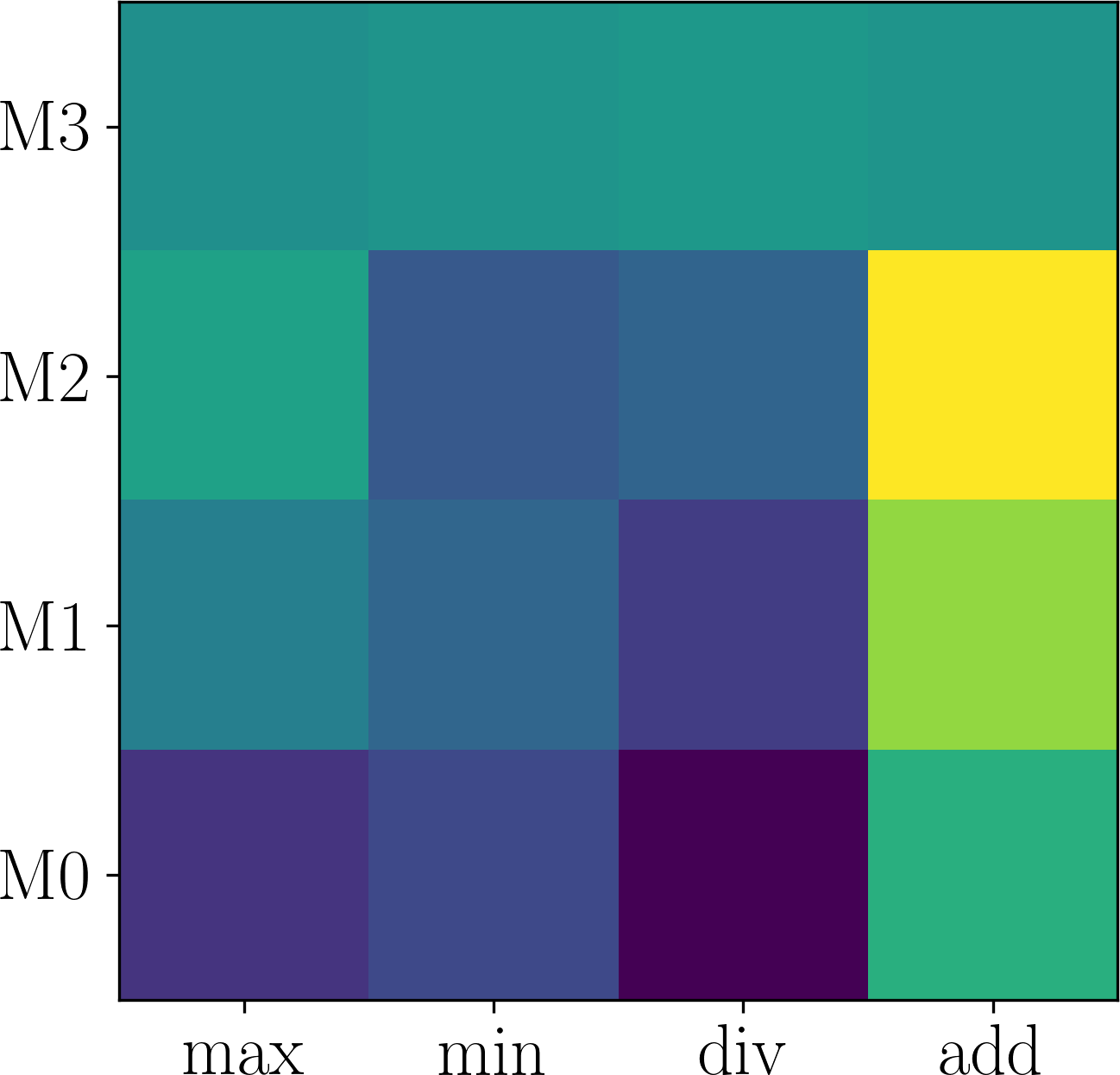}}
	\subfigure[Permutated.]{
		\includegraphics[width=0.174\textwidth]{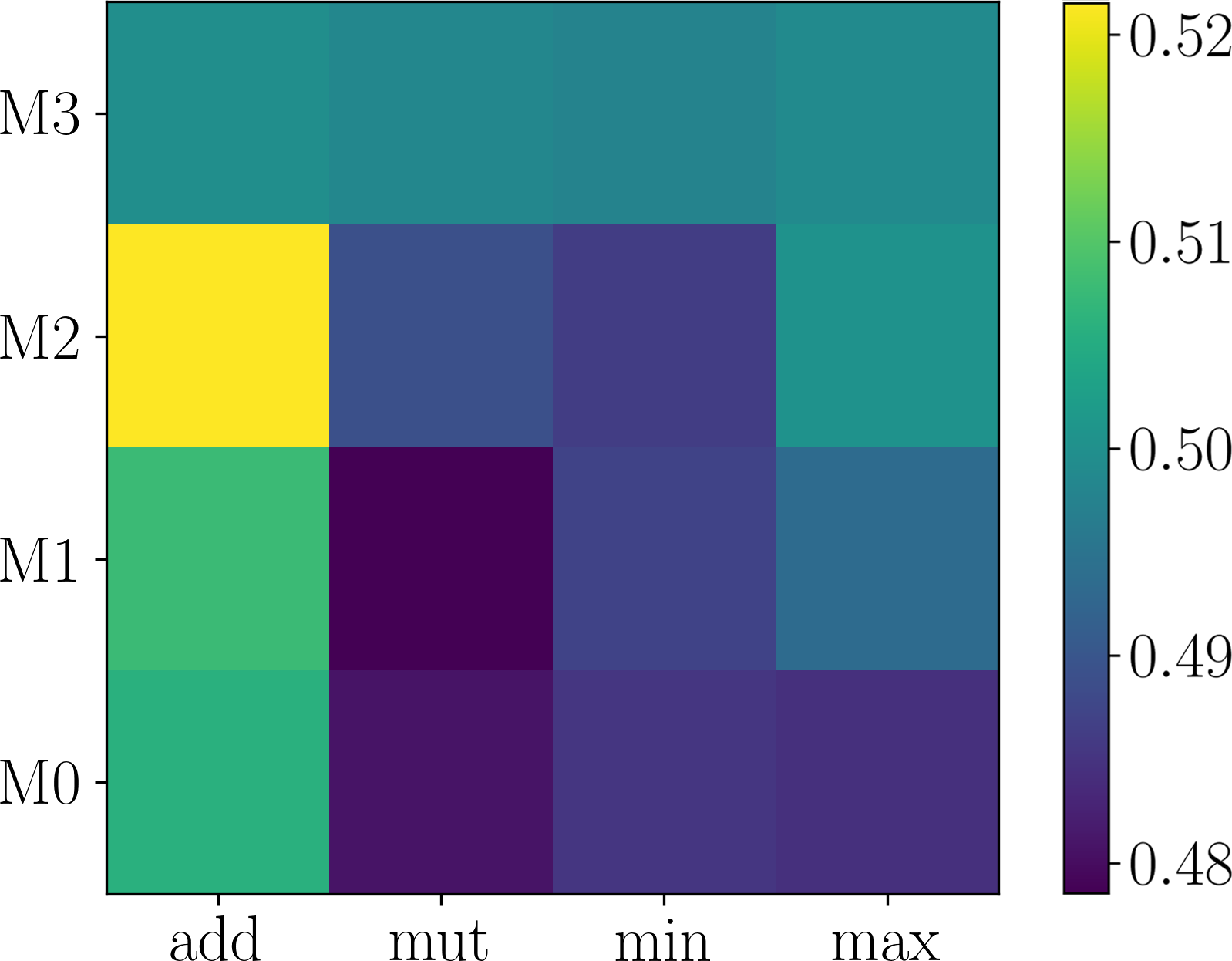}}
	\vspace{-10px}
	\caption{Visualization of the $\{\alpha^{l,k}\}$ obtained on Last.fm, where operations and orders are changed.	
	}
	\label{fig:prop-property}
	\vspace{-10px}
\end{figure}

Further, 
we validate the inner-group consistence and permutation-invariance properties proved in Proposition~\ref{prop:space}. 
Comparing Figure~\ref{fig:prop-assump} (a) with Figure~\ref{fig:prop-property} (a-b), one can see that
changing $\odot$ to $/$ and $+$ to $-$ obtains the same results, which validates inner-group consistency.  
Finally, comparing Figure~\ref{fig:prop-assump} (a) with Figure~\ref{fig:prop-property} (c), one can observe that inter-group permutation invariant also holds. 

\subsection{Sensitivity Analysis  (RQ3)} 
\label{sec:exp:study}

Finally, we conduct sensitivity analysis of \TheName{} on 
Last.fm. 

\subsubsection*{Effects of $K$.}
Recall that we keep modulation operations corresponding to the Top-$K$ largest $\alpha^{l,k}$s in modulation structure. 
Figure~\ref{fig:last-k} plots the effect of changing $K$. As shown, $K$ cannot be too small that the model is 
unable to capture enough user-specific preference, 
nor too large that the model can overfit to the limited interaction history of cold-start users. 

\subsubsection*{Effect of the Depth of Predictor.}
Figure~\ref{fig:last-mlp} plots the effect of using predictors with different $L$ number of layers. 
One can observe that choosing different $L$ in a certain range has low impact on the performance, while choosing $L=4$ is already good enough to obtain the state-of-the-art results as shown in Table~\ref{tab:results}. 

\subsubsection*{Effects of Support Set Size.}
Figure~\ref{fig:last-support} shows the performance of users with different length of history. 
During inference, we randomly sample only a portion of interactions from the original support set $\cS_i$ of each $T_i$ in $T^\text{test}$. 
As can be seen,  prediction would be more accurate given interaction history, which may alleviate the bias in representing the user-specific preference.

%
%
%

\section{Conclusion}
We propose a modulation framework called \TheName{} for user cold-start recommendation.  
In particular, we use a hypernetwork to map each user's history interactions to user-specific parameters, 
which are then used to modulate the predictor. 
We design a search space for modulation functions and positions, 
which not only covers existing modulation-based models but also has the ability to find more effective structures. 
We theoretically prove the space could be transformed to a smaller space, where we can search for modulation structure efficiently and robustly. 
Extensive experiments show that \TheName{} performs the best on benchmark datasets. 
Besides, \TheName{} can efficiently find proper modulation structure for different data, which make it easy to be deployed in recommendation systems. 

\section*{Acknowledgment}
Q. Yao was in part supported by NSFC (No. 92270106) and CCF-Baidu Open Fund.


\bibliographystyle{ACM-Reference-Format}
\bibliography{coldnas}

\appendix

\clearpage
\section{Details of Model Structure}
\label{app:structure}
As mentioned in Section~\ref{sec:proposed-space}, existing models ~\cite{dong2020mamo,lin2021task,pang2022pnmta} share an architecture consisting of three components: embedding layer $E$, adaptation network $A$, and predictor $P$. 
Here we provide the details of these components used in \TheName{}. 

\subsection{Embedding Layer}
We first embed user and item’s one-hot categorical features into dense vectors through the embedding layer. 
Taking $u_i$ for example, we generate a content embedding for each categorical content feature and concatenate them together to obtain the initial user embedding. 
Given $B$ user contents, the embedding is defined as:
\begin{align}\label{eq:emb}
	\bm{u_i}=[~\bm{W}^1_E\bm{c}_i^1\mid{\bm{W}^2_E\bm{c}_i^2}\mid\cdots\mid{\bm{W}^B_E\bm{c}_i^B}~]
\end{align}
where $[~\cdot\mid\cdot~]$ is the concatenation operation, $\bm{c}_i^b$  is the one-hot vector of the $b$th categorical content of $u_i$ , and $\bm{W}^b_E$ represents the embedding matrix of the corresponding feature in the shared user feature space. 
The parameters of embedding layer are collectively denoted as $\bm{\theta}_E=\{\bm{W}^b_E\}_{b=1}^B$. 

\subsection{Adaptation Network}
\label{app:adapte}

We first use a fully connected (FC) layer 
to get hidden representation of interaction history $\bm{r}_{i,j}$, which is calculated as 
\begin{equation}\label{eq:adp}
	\bm{r}_{i,j}=\text{ReLU}(\bm{W}^1_{A}[~\bm{u}_{i}\mid \bm{v}_{j} \mid y_{i,j}~]+\bm{b}^1_{A}). 
\end{equation}
We use mean pooling to aggregate the interactions in $\cS_i$ as the task context information of $u_i$, i.e.,preference,
$\bm{c}_i=\frac{1}{N}\sum_{1}^{N}\bm{r}_{i,j}$.
Then, we use another FC layer to generate the user-specific parameters as 
$\bm{\phi}_i=\bm{W}^2_{A}\bm{c}_i+\bm{b}^2_{A}$. 
The parameters of adaptation network is denoted as $\bm{\theta}_{A}=\{\bm{W}^1_{A},\bm{b}^1_{A},\bm{W}^2_{A},\bm{b}^2_{A}\}$.

\subsection{Predictor}
We use a $L$-layer MLP as the predictor. 
Denote output of the $l$th layer hidden units as $\bm{h}^l$, the modulation function at the $l$th layer as $M^l$, we obtain prediction by
\begin{align}\nonumber
	\bm{h}^l_i
	& = M_t(\bm{h}^l, \bm{\Phi}_i ),
	\\\label{eq:mlp}
	\bm{h}^{l+1}
	& = \text{ReLU}( \bm{W}^l_P \bm{h}_i^l + \bm{b}^l_P),
\end{align} 
where $l\in[0,1,\cdots L-1]$, $\bm{h}^0=[\bm{u}_{i}\mid \bm{v}_{j}]$ and $\hat{y}_{i,j}=\bm{h}^L$. The parameters of predictor are denoted as $\bm{\theta}_{P}=\{\bm{W}^l_P,\bm{b}^l_P\}_{l=0}^{L-1}$.

\section{Proof of the Proposition~\ref{prop:space}}
\label{app:proof}

\begin{proof}
	To prove Proposition~\ref{prop:space}, we need to use the two conditions below:
	
	\noindent
	\textbf{Condition 1:} 
	If $\phi_i^k$ is the input of operation $\odot$ or $/$, then every element in $\phi_i^k$ is non-negative. We implement this as ReLU function.
	
	\noindent
	\textbf{Condition 2:} The adaptation network $A$ is expressive enough, that it can approximate 
	$\hat{\phi}_i = \phi^p_i\circ_{\text{op}^1}\phi^q_i$, 
	where $\circ_{\text{op}^1}\in \mathcal{O}$ and $\phi^p_i$, $\phi^q_i \in \bm{\Phi}_i$ 
	learned from the data. This naturally holds due to the assumption.
	
	Divide $\circ_{\text{op}}$ into 4 groups: $G_1=\{\max\},~G_2=\{\min\},~G_3=\{+,-\},~G_4=\{\odot,/\}$.
	We can prove two important properties below. 
	
	\paragraph{Property 1: Inner-group consistence}.We can choose an operation in each group as the group operation $\circ_{g_i}$, e.g. $\circ_{g_1}=\max,~\circ_{g_2}=\min,~\circ_{g_3}=+,~\circ_{g_4}=\odot$. Then any number of successive operations that belong to the same group $G_i$ can be associated into one operation $\circ_{g_i}$, i.e., 
	$$\bm{x}\circ_{\text{op}^k} \bm{\phi}^k_i \circ_{\text{op}^{k+1}}\bm{\phi}^{k+1}_i \cdots \circ_{\text{op}^{k+m}} \bm{\phi}^{k+m}_i=\bm{x}\circ_{g_i}\hat{\bm{\phi}}_i^k, $$
	$$s.t.~ \circ_{\text{op}^k}, \circ_{\text{op}^{k+1}}\cdots \circ_{\text{op}^{k+m}} \in G_i$$
	It's trivial to prove with condition 2, e.g.,
	$\bm{x}+\bm{\phi}_i^1-\bm{\phi}_i^2+\bm{\phi}_i^3=\bm{x}+\hat{\bm{\phi}}_i^1$, 
	where $\hat{\bm{\phi}}_i^1=\bm{\phi}_i^1-\bm{\phi}_i^2+\bm{\phi}_i^3$.
	
	\paragraph{Property 2: Inter-group permutation-invariance.} The operations in different groups are permutation-invariant, i.e.,
	$$\bm{x}\circ_{g_i}\bm{\phi}^k_i \circ_{g_j}\bm{\phi}^{k+1}_i =\bm{x}\circ_{g_j}\hat{\bm{\phi}}^k_i\circ_{g_i}\hat{\bm{\phi}}^{k+1}_i$$
	It is also trivial to prove with condition 1 and condition 2, e.g.,
	$$\max(\bm{x},~\bm{\phi}_i^1)+\bm{\phi}_i^2=\max(\bm{x}+\hat{\bm{\phi}}_i^1,~\hat{\bm{\phi}}_i^2),$$ where $\hat{\bm{\phi}}_i^1=\bm{\phi}_i^2$ and $\hat{\bm{\phi}}_i^2=\bm{\phi}_i^1+\bm{\phi}_i^2$.
	
	With the above two properties, we can recurrently commute operations until operations in the same group are gathered, and merge operations in the four groups respectively. So 
	$$\bm{h}^l_i 
	= \bm{h}^l \circ_{\text{op}^1} \bm{\phi}_i^1 \circ_{\text{op}^2} \bm{\phi}_i^2\cdots \circ_{\text{op}^C} \bm{\phi}_i^C $$equals to$$
	\bm{h}^l_i = \bm{h}^l \circ_{g_1} \hat{\bm{\phi}}_i^1 \circ_{g_2} \hat{\bm{\phi}}_i^2\circ_{g_3} \hat{\bm{\phi}}_i^3 \circ_{g_4} \hat{\bm{\phi}}_i^4,$$ and the four group operations are permutation-invariant. 
	
	Note that since the identity element (\textbf{0} or \textbf{1}) of each group operation can be learned from condition 2, a modulation function that doesn't cover all groups can also be represented in the same form.
\end{proof}

\section{Experiments}
\label{app:expt}

\subsection{Experiment Setting}
\label{app:expt-setting}

Experiments were conducted on  a 24GB NVIDIA GeForce RTX 3090 GPU, with Python 3.7.0, CUDA version 11.6. 
\subsubsection{Hyperparameter Setting}
We find hyperparameters using the $\cT^{\text{val}}$ via grid search for existing methods. 
In \TheName{}, the batch size is 32, $\bm{W}^b_E$ in \eqref{eq:emb} has size $32\times|c_i^b|$ where $|c_i^b|$ is the length of $c_i^b$. 
The dimension of hidden units in \eqref{eq:mlp}  is set as $\bm{h}^{1}=128, \bm{h}^{2}=64, \bm{h}^{3}=32$ for all three datasets. 
The learning rate $\beta$ is chosen from $\{5\times 10^{-6},1\times 10^{-5},5\times 10^{-5},1\times 10^{-4}\}$  and 
the dimension of $\bm{r}_{i,j}$ in \eqref{eq:adp} is chosen from $\{128,256,512,1024\}$. 
The final hyperparameters chosen on the three benchmark datasets are shown in Table~\ref{tab:hyperpara}. 

\begin{table}[htbp]
	\centering
	\setlength\tabcolsep{3pt}
	\caption{The final hyperparameters chosen in \TheName{}.}
	\resizebox{0.47\textwidth}{!}{%
		\begin{tabular}{c |c | c|c}
			\toprule
			Dataset&	{MovieLens}  &  {BookCrossing}  & 	{Last.fm}\\ \midrule
			$\beta$&$5\times 10^{-5}$&$5\times 10^{-6}$&$1\times 10^{-4}$\\ \midrule
			$\bm{r}_{i,j}$&1024&512&256\\ \midrule
			$\bm{W}^1_{A}$ &$1024\times 289$&$512\times 161$&$256\times 65$\\ \midrule
			$\bm{W}^2_{A}$ &$2052\times 1024$&$1540\times 512$&$1156\times 256$\\ \midrule
			$\bm{b}^1_{A}$&$1024$&$512$&$256$\\ \midrule
			$\bm{b}^2_{A}$&$2052$&$1540$&$1156$\\ \midrule
			$\bm{W}^0_{P}$&$128\times 289$&$128\times 161$&$128\times 65$\\\midrule 
			$\bm{W}^1_{P}$&$64\times 128$&$64\times 128$&$64\times 128$\\\midrule 
			$\bm{W}^2_{P}$&$32\times 64$&$32\times 64$&$32\times 64$\\\midrule 
			$\bm{W}^3_{P}$&$1\times 32$& $1\times 32$&$1\times 32$\\\midrule 
			$\bm{b}^0_{P}$&$128$&$128$&128\\\midrule 
			$\bm{b}^1_{P}$&$64$&$64$&64\\\midrule 
			$\bm{b}^2_{P}$&$32$&32&32\\\midrule 
			$\bm{b}^3_{P}$&1&1&1\\\midrule 
			maximum iteration number&50&50&100\\\bottomrule
		\end{tabular}
	}
	\label{tab:hyperpara}
\end{table}

\begin{figure*}[htbp]
	\centering
	\vspace{-10px}
	\subfigure[Varying $K$ in Top-$K$.]{
		\includegraphics[width=0.32\textwidth]{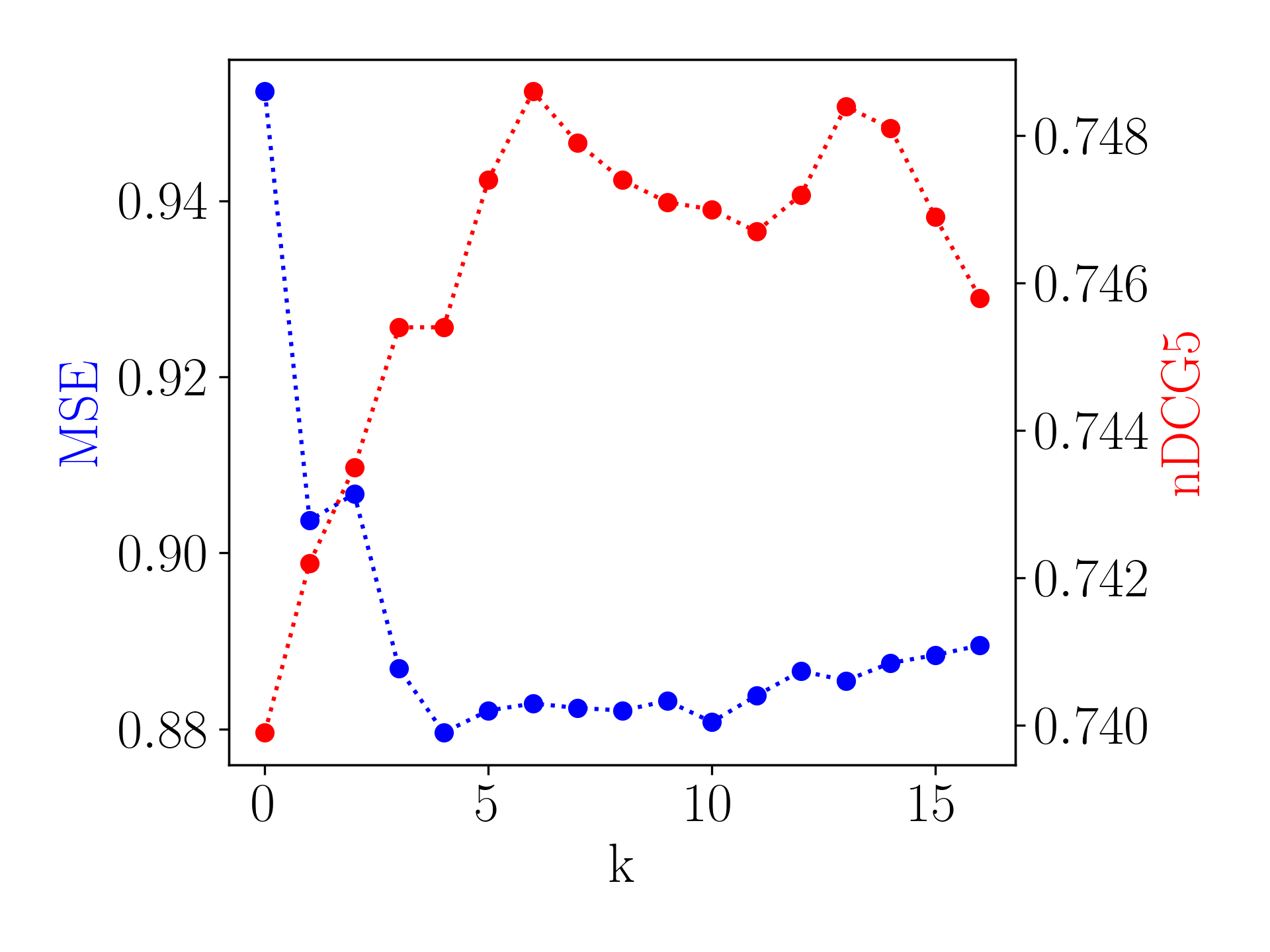}}
	\subfigure[Varying number of layers of the predictor.]{
		\includegraphics[width=0.32\textwidth]{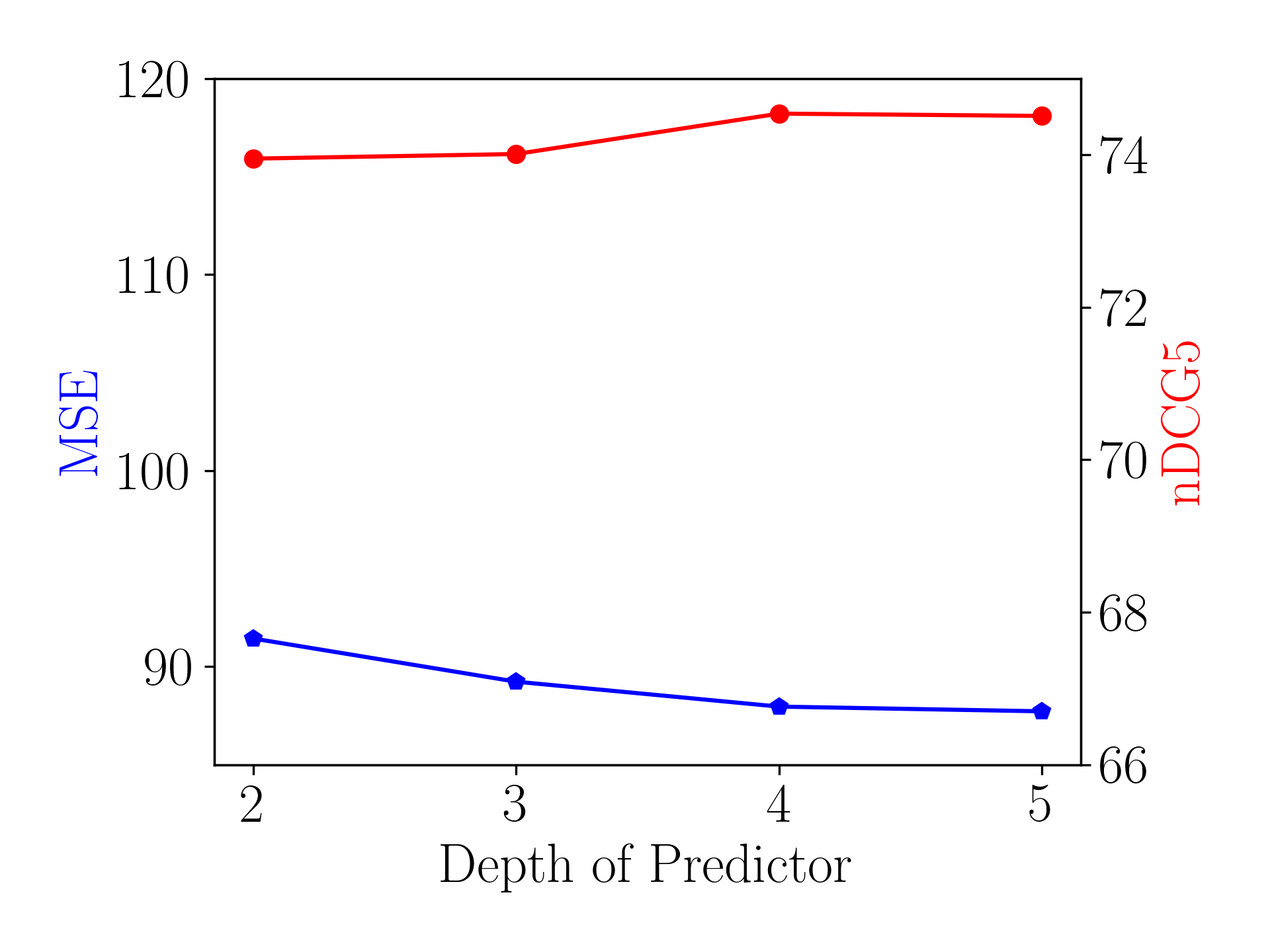}}
	\subfigure[Varying support set size during inference.]{
		\includegraphics[width=0.32\textwidth]{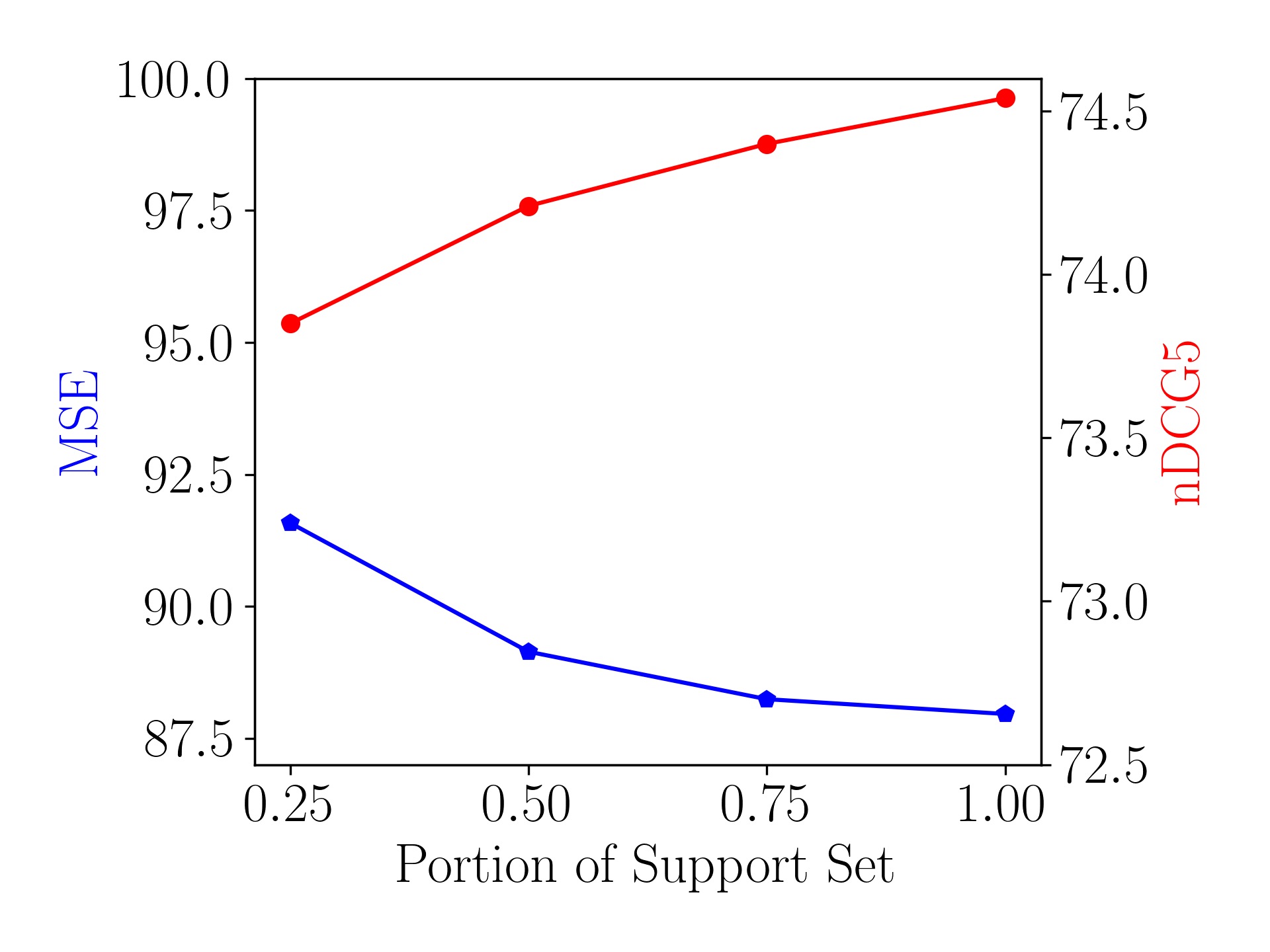}}
	\vspace{-10px}
	\caption{Model sensitivity analysis on MovieLens.}
	\label{fig:sensitivity-movielens}
	\vspace{-8px}
\end{figure*}

\begin{figure*}[htbp]
	\centering
	\subfigure[Varying $K$ in Top-$K$.]{
		\includegraphics[width=0.32\textwidth]{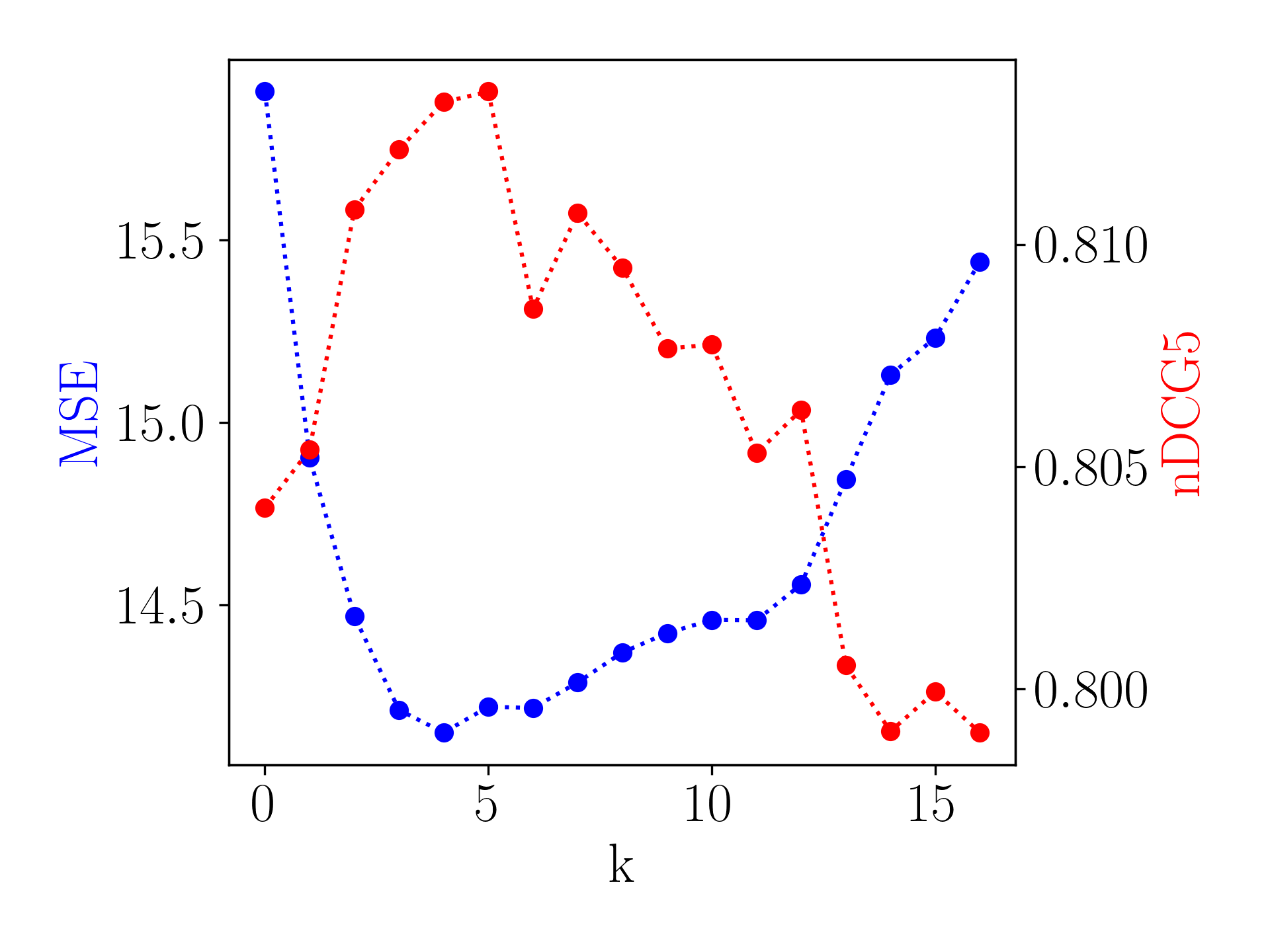}}
	\subfigure[Varying number of layers of the predictor.]{
		\includegraphics[width=0.32\textwidth]{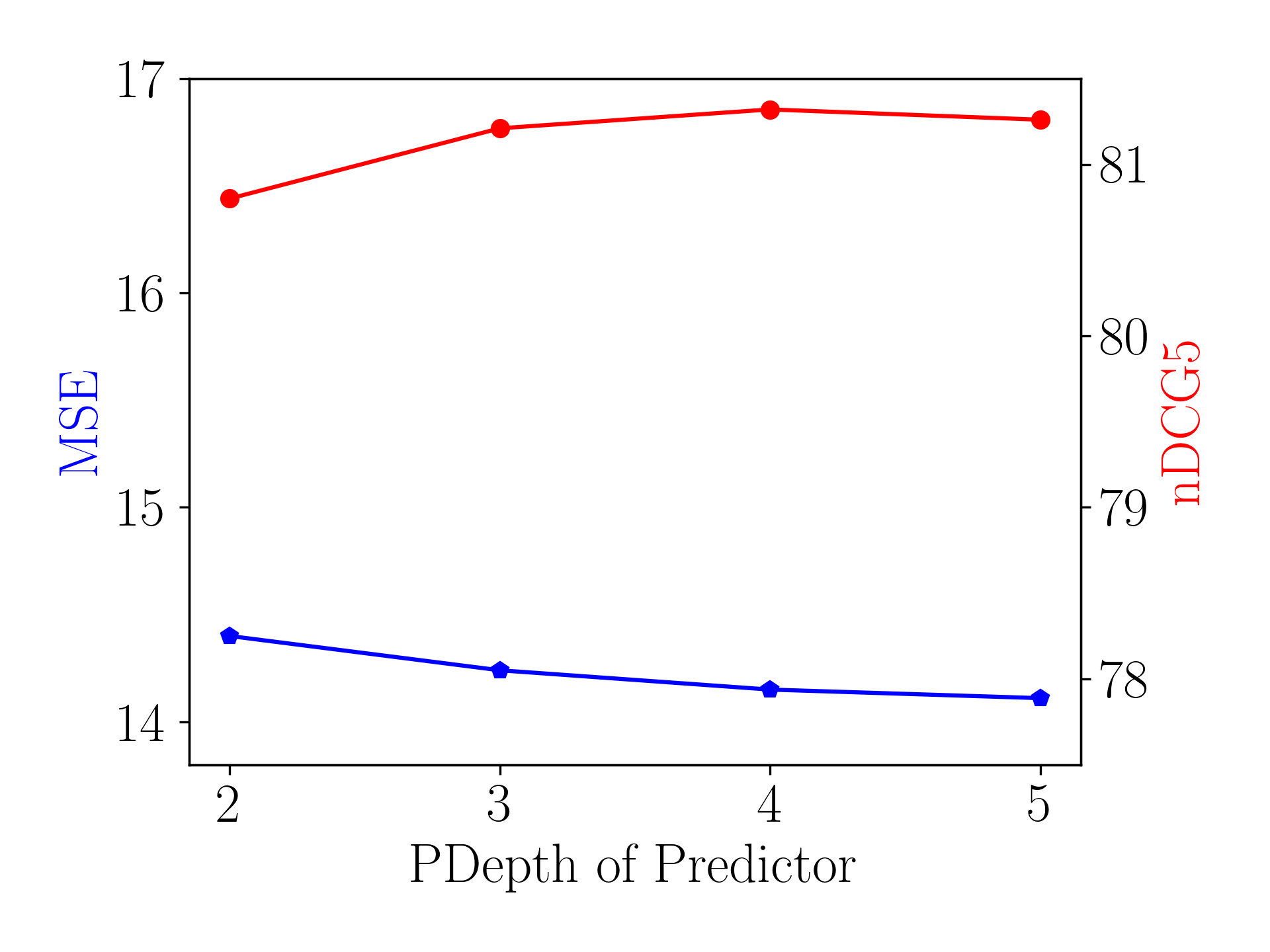}}
	\subfigure[Varying support set size during inference.]{
		\includegraphics[width=0.32\textwidth]{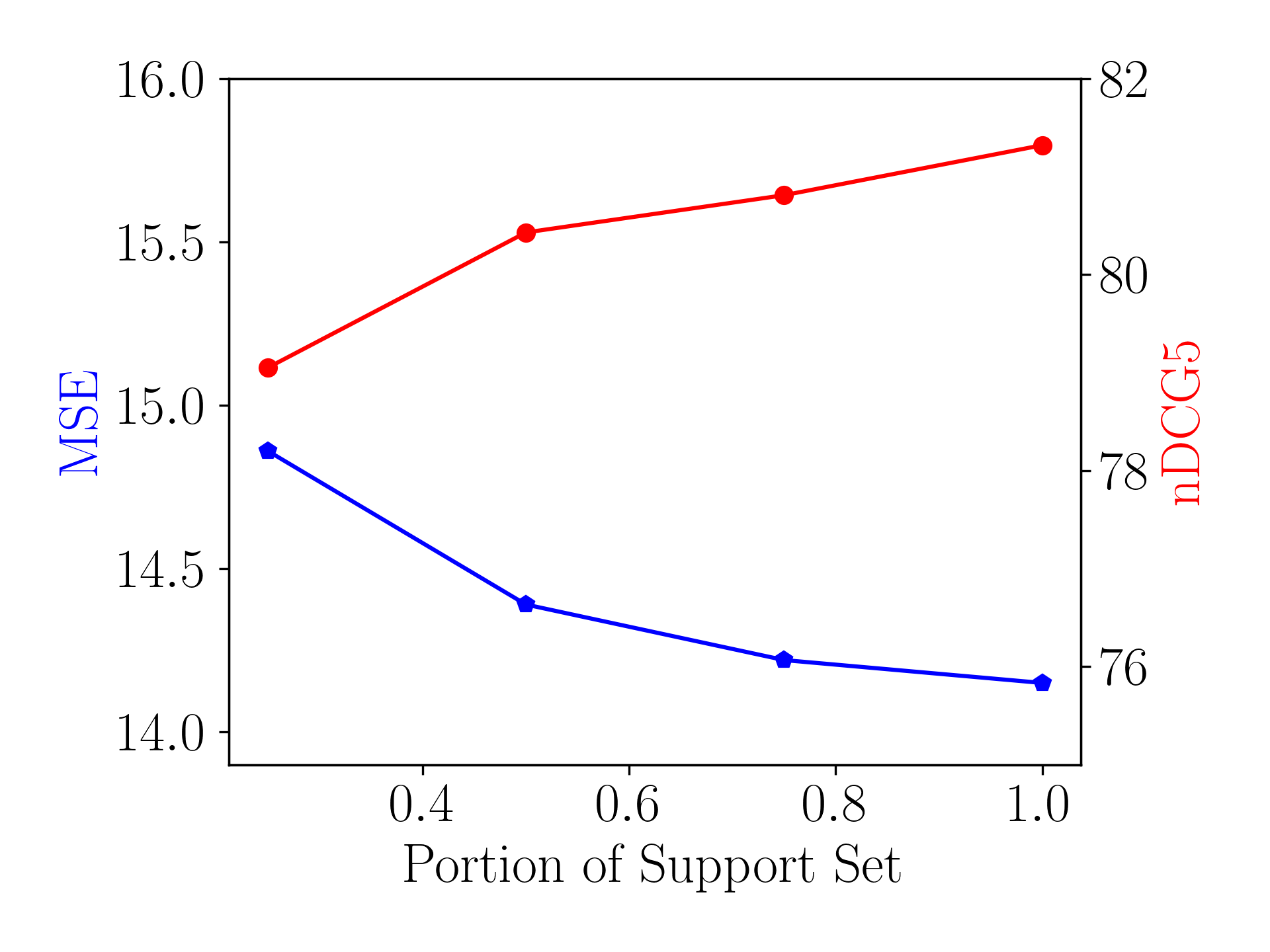}}
	
	\vspace{-10px}
	\caption{Model sensitivity analysis on BookCrossing.}
	\label{fig:sensitivity-book}
	\vspace{-8px}
\end{figure*}



\subsubsection{URLs of Datasets and Baselines}

We use three benchmark datasets (Table~\ref{tab:data-stat}):  
(i) \textbf{MovieLens}\footnote{\url{https://grouplens.org/datasets/movielens/1m/}}~\cite{harper2015movielens}: a dataset containing 1 million movie ratings of users collected from MovieLens, whose features include gender, age, occupation, Zip code, publication year, rate, genre, director and actor; 
(ii) \textbf{BookCrossing}\footnote{\url{http://www2.informatik.uni-freiburg.de/~cziegler/BX/}}~\cite{ziegler2005improving}: a collection of users' ratings on books in BookCrossing community, whose features include age, location, publish year, author, and publisher; 
and 
(iii) \textbf{Last.fm}\footnote{\url{https://grouplens.org/datasets/hetrec-2011/}}: a collection of user's listening count of artists from Last.fm online system, whose features only consist of user and item IDs. 
In experiments, we compare \textbf{\TheName{}} with the following representative user cold-start methods: 
(i)
traditional deep cold-start model 
\textbf{DropoutNet}\footnote{\url{https://github.com/layer6ai-labs/DropoutNet}}~\cite{volkovs2017dropoutnet} 
and
(ii) FSL based methods include
\textbf{MeLU}\footnote{\url{https://github.com/hoyeoplee/MeLU}}~\cite{lee2019melu},
\textbf{MetaCS} \cite{bharadhwaj2019meta}, 
\textbf{MetaHIN}\footnote{\url{https://github.com/rootlu/MetaHIN}}~\cite{lu2020meta}, 
\textbf{MAMO}\footnote{\url{https://github.com/dongmanqing/Code-for-MAMO}}~\cite{dong2020mamo}, 
and 
\textbf{TaNP}\footnote{\url{https://github.com/IIEdm/TaNP}}~\cite{lin2021task}.  
MetaCS is very similar to MeLU, except that it updates all parameters during meta-learning. Hence, we implement MetaCS based on the codes of MeLU.

\subsection{More Experimental Results}
\label{app:expt-results}

Here, we show empirical results of Section \ref{sec:exp:study} of MovieLens and BookCrossing in Figure \ref{fig:sensitivity-movielens} and \ref{fig:sensitivity-book}. The results of all three datasets show similar patterns and the same analysis and conclusions in Section \ref{sec:exp:study} are applicable.

\subsection{Complexity Analysis}
Denote the layer number of adaptation network as $L_A$, the layer number of predictor as $L_P$, the support set size as $N$, and the query set size as $M$. For notation simplicity, we denote the hidden size of each layer is the same as $D$.
In the search phase (Algorithm \ref{alg:coldnas} step 2$\sim$10), the calculation of $\bm{\Phi}_i$ of a task is $O(NL_AD^3)$, and predicting each item costs $O(L_PD^3)$, so the average complexity is $O((L_P+\frac{NL_A}{M})D^3)$. Similarly, in the retrain  (Algorithm \ref{alg:coldnas} step 12$\sim$13), the time complexity is also $O((L_P+\frac{NL_A}{M})D^3)$. In total, the time complexity is $O((L_P+\frac{NL_A}{M})D^3)$.

\end{document}